\documentclass{article}

\usepackage{microtype}
\usepackage{graphicx}
\usepackage{subfigure}
\usepackage{booktabs} 
\usepackage{amsfonts}
\usepackage{amsmath}
\usepackage{enumitem}
\usepackage{multirow}
\usepackage{pgfplots}
\usepackage{tablefootnote}
\usepackage{multicol}
\pgfplotsset{width=0.88\columnwidth,compat=1.9}
\usepackage{hyperref}

\DeclareMathOperator*{\argmin}{arg\,min}
\usepackage[accepted]{mlsys2024}

\mlsystitlerunning{Decoupled Vertical Federated Learning}

\begin{document}

\twocolumn[
\mlsystitle{Decoupled Vertical Federated Learning \\ for Practical Training on Vertically Partitioned Data}


\mlsyssetsymbol{equal}{*}

\begin{mlsysauthorlist}
\mlsysauthor{Avi ~Amalanshu}{to}
\mlsysauthor{Yash Sirvi}{to}
\mlsysauthor{David I. Inouye}{goo}
\end{mlsysauthorlist}

\mlsysaffiliation{to}{Indian Institute of Technology Kharagpur, Kharagpur, WB, India }
\mlsysaffiliation{goo}{Purdue University West Lafayette, IN, USA}

\mlsyscorrespondingauthor{Avi ~Amalanshu}{avi.amalanshu@kgpian.iitkgp.ac.in}

\mlsyskeywords{Machine Learning, MLSys}

\vskip 0.3in

\begin{abstract}
    Vertical Federated Learning (VFL) is an emergent distributed machine learning paradigm for collaborative learning between clients who have disjoint features of common entities. However, standard VFL lacks fault tolerance, with each participant and connection being a single point of failure. Prior attempts to induce fault tolerance in VFL focus on the scenario of ``straggling clients", usually entailing that all messages eventually arrive or that there is an upper bound on the number of late messages. To handle the more general problem of arbitrary crashes, we propose Decoupled VFL (DVFL). To handle training with faults, DVFL decouples training between communication rounds using local unsupervised objectives. By further decoupling label supervision from aggregation, DVFL also enables redundant aggregators. As secondary benefits, DVFL can enhance data efficiency and provides immunity against gradient-based attacks. In this work, we implement DVFL for split neural networks with a self-supervised autoencoder loss. When there are faults, DVFL outperforms the best VFL-based alternative (97.58\% vs 96.95\% on an MNIST task). Even under perfect conditions, performance is comparable.
\end{abstract}
]



\printAffiliationsAndNotice{\mlsysEqualContribution} 

\section{Introduction}
\label{Introduction}
Federated Learning \citep{pmlr-v54-mcmahan17a}, or FL, was introduced by Google researchers as a strategy for distributed learning, addressing communication efficiency and data privacy. Distributed participants in FL training do not expose their data to any other party. Instead, they train local models on their data, guided by a coordinator that has no knowledge of the agents' data. Thus, FL extends collaborative machine learning to contexts where data communication is undesirable or forbidden. Recent advances in communication systems and cloud computing infrastructure combined with an increased public awareness and legislation regarding data privacy (such as the EU's GDPR) have contributed to FL's advancement as an emergent machine learning paradigm. 

Among FL's developments is Vertical Federated Learning (VFL). Standard FL can be said to have a ``horizontal" or “sample-parallel” division of data: each participant holds unique samples within a shared feature space. Conversely, VFL participants hold unique features of a common sample space, which is a ``vertical", ``feature-parallel” division. This allows guest agents who individually have incomplete and information about their target to learn meaningful joint representations without sharing data.

VFL has received nascent interest, with some real-world proposals and applications \citep{liu2023vertical}. However, VFL requires careful synchronization which makes it difficult to engineer and scale. Critically, training is contingent on the exchange of intermediate results between the guests and the host. A crashed connection or participant on the backward pass means a missed model update. On the forward pass, it is catastrophic: an unfulfilled model input. 

We propose Decoupled Vertical Federated Learning (DVFL), where guests and hosts each train asynchronously on their own unsupervised objective before the label owner learns a transfer learning model from its labels. At an algorithmic level, DVFL avoids the forward and backward locking between hosts and guests. Hence, it is free of a single point of failure. Such ``localized" training means that other participants can continue training even when one fails. Further, separating label inference from aggregation allows for redundant aggregators, further localizing faults.

The exchange of intermediate results is also the root of many other limitations of VFL including \emph{inference attacks}. Feedback received from the host has mutual information with labels and other guests' features, which can be exploited by curious guests (mitigating this is typically costly in computation and performance: see \citet{yang2023survey,khan2023vertical}). Another is the inability to learn from features pertaining to samples outside the intersection of all participants' sample spaces. DVFL does not suffer from these problems.

Altogether, we expect DVFL to enable learning from vertically partitioned data in scenarios where they were so far infeasible, and help scale implementations where they weren't. 
The main contributions of this work are:
\begin{enumerate}
    \item We present DVFL, a strategy that eschews both host-to-guest and label-to-host feedbacks, and therefore enables redundancy, has inbuilt fault handling (\emph{including those during the backward pass}) and inherent gradient privacy, and performs comparably to VFL.
    \item We show host redundancy improves model performance, especially when there are faults.
    \item We demonstrate how DVFL guests might exploit (possibly unlabeled) data outside a limited sample intersection to learn strong representations.
\end{enumerate}

\if Among FL's developments is Vertical Federated Learning (VFL). Standard FL can be said to have a ``horizontal" or “sample-parallel” division of data: each participant holds unique samples within a shared feature space. Conversely, VFL participants hold unique features of a common sample space, which is a ``vertical", ``feature-parallel” division. This allows guest agents who individually have incomplete information about their target to learn joint representations without sharing data. 

VFL has received nascent interest, with some real-world proposals and applications \citep{liu2023vertical}. However, VFL suffers from flaws that make its implementation impractical even in highly controlled cross-silo environments, and altogether infeasible otherwise. Critically, the link between the guests and the host is a single point of failure. Training is contingent on the exchange of intermediate results between the guests and the host. A crashed connection or participant on the backward pass means a missed model update. On the forward pass, it is catastrophic: an unfulfilled model input.

In fact, this exchange of intermediate results is also the root of many other limitations of VFL. In order to make a meaningful prediction, all guests must be processing the features of the same agreed upon entity at any time. Entities outside the intersection of all participants cannot be used for training. Another example is the risk of \emph{inference attacks} from curious guests, since feedback received from the host has mutual information with labels and other guests' features. Mechanisms to mitigate inference attacks are expensive and hamper model convergence \citep{yang2023survey,khan2023vertical}. 

We propose Decoupled Vertical Federated Learning (DVFL), where guests and hosts each train on their own unsupervised objective before the label owner learns a transfer learning model from its labels. At an algorithmic level, DVFL avoids the forward and backward locking between hosts and guests. Hence, it is free of a single point of failure. Since there is no feedback across participants, it is immune to inference attacks. This approach also yields two useful system properties. One, this allows redundant hosts. Standard VFL would require hosts to share labels with each other, which is forbidden. Two, participants train independently and asynchronously to each other. Since guests train independently of each other, we may use separate datasets for inference and representation learning. That is to say, we may train guest and host models on data which is not necessarily labeled and not necessarily a part of a joint sample space.

The main contributions of this work are:
\begin{enumerate}
    \item We present DVFL, a strategy that eschews both host-to-guest and label-to-host feedbacks, and therefore enables redundancy, has inbuilt fault handling (\emph{including those during the backward pass}) and inherent gradient privacy, and performs comparably to VFL.
    \item We show host redundancy improves model performance, especially when there are faults.
    \item We demonstrate how DVFL guests might exploit (possibly unlabeled) data outside a limited sample intersection to learn strong representations.
\end{enumerate}
\fi
\section{Background: Standard VFL}
\label{sec:vfl}
\subsection{Notation}
The rows of a dataset are features of a unique entity $\mathbf x$. We may serialize these entities using a sample index e.g. as $\mathbf x_j$. $g_i$ is the $i^{\text{th}}$ data-owning agent, member of set of guests $\mathcal G$. $\mathcal F_i$ is the set of features that guest $g_i$ records. $\mathbf x_{j,\mathcal F_i}$ are the features $\mathcal F_i$ of the $j^\text{th}$ entity. $m_i(\text{ }\cdot\text{ };\theta_i)$ is $g_i$'s model, $\mathbb R^{|\mathcal F_i|}\to \mathbb R^{k_i}$ function parameterized by $\theta_i$, which is learnable. $\mathcal S_i$ is the set of sample indices for which $g_i$ has records. $\mathcal S_\text{guests}$ = $\bigcap_{i = 1}^{|\mathcal G|}$. $h$ is the host agent, which owns labels. $\mathcal S_\text{labels}$ is the set of sample indices corresponding to labeled entities. $m_h(\text{ }\cdot\text{ }; \theta_h)$ is the host's $\mathbb R^{k_1}\times\mathbb R^{k_2}\dots\times\mathbb R^{k_{|\mathcal G|}}\to\mathbb R^{\mathrm{out}}$ learnable model, which accepts the outputs of $m_i$ and makes a task-relevant prediction.

\subsection{The VFL Hierarchy}
\label{sec:hierarchy}
\textbf{Guest}: The guest role does not entail access to labels or other guests' data. The guest's private model $m_i$'s task is to concisely represent the features $\mathcal F_i$ of $\mathbf x_j$ $\forall$ sample indices $j\in\mathcal S_\text{guests}\cap\mathcal S_\text{labels}$. Guests obtain these encodings and pass them to the host $h$.

\textbf{Host:} The host agent $h$ 
executes every round of training. The host has access to target labels. The host, given representations for all available features of $\mathbf x_j$, may use its own model to make useful predictions for $\mathbf x_j$. 

If the host has access to the full computational graph, it may adjust all parameters to better predict label $\mathbf y$, usually by minimizing the expected value of some loss function. Otherwise, it may update only its parameters $\theta_h$ and send intermediate results (e.g. gradients) via which the guests may update their own models. 

Together, the guests and host solve the following joint optimization problem:
$$\theta^* = \argmin_{\theta}\mathbb E_{\mathbf x_j\sim X}\left[\ell(m_h\left(\hat{\mathbf x}_{j,1},\dots,\hat{\mathbf x}_{j,{|\mathcal G|}}); \theta_h\right), \mathbf y_j)\right]\,,$$ 
where $\hat {\mathbf x}_{ j,i}\equiv m_i(\mathbf x_{j, \mathcal F_i}; \theta_i) \in \mathbb R^{k_i}$.

It is possible that the host may also be one of the guests. In such situations, the host is often known as the ``active party" and guests as ``passive parties."

\subsection{Split Training for Neural Networks}
\begin{figure*}
    \centering
    \includegraphics[width=0.9\textwidth]{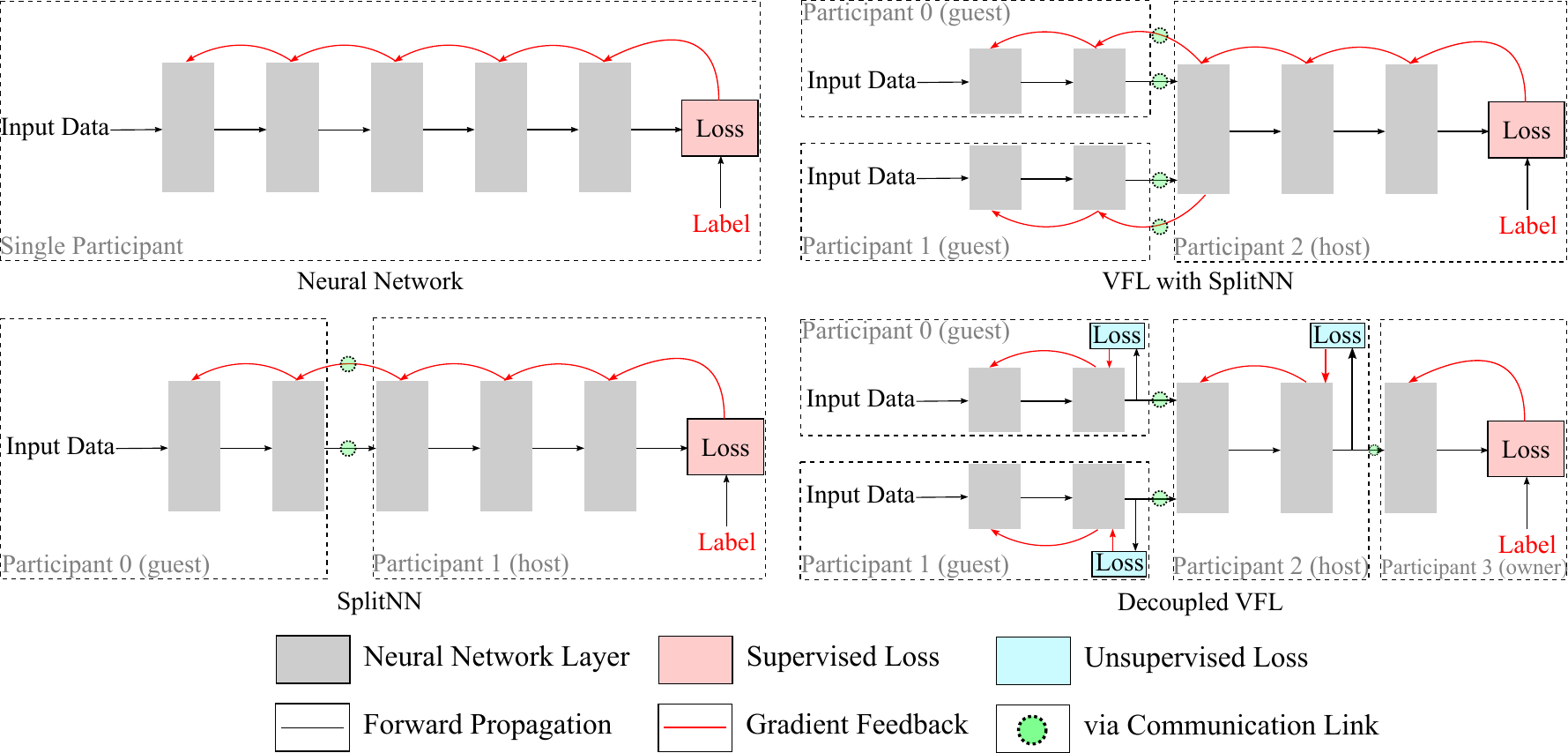}
    \caption{Diagram illustrating the distributed training of a neural network under SplitNN, VFL with SplitNN, and DVFL. The two input data for VFL and DVFL are partial features of the same entity.}
    \label{fig:SplitNN}
\end{figure*}

SplitNN \citep{Gupta2018} is a faithful implementation of VFL for NNs. It was originally proposed as an algorithm to distribute the training of a global neural network across two agents by splitting the network depth-wise into two. At the ``split layer", forward and backward signals are communicated between the two devices.\footnote{One might say that the standard SplitNN addresses a special case of VFL where the host has access only to labels and the (single) guest has access only to features. The original paper \citep{Gupta2018} also presents a protocol for training when the features are horizontally distributed across $N$ guests.} \citet{ceballos2020splitnndriven, PyVertical} extend SplitNN for vertically partitioned data by vertically splitting the shallower layers (see Figure \ref{fig:SplitNN}). 
Effectively, the guest models $m_i$ are NNs and host model $m_h$ consists of a concatenation layer $m_h^0$ followed by an NN $m_{h}^1$.

\subsection{Impracticalities in VFL Training}
VFL is contingent on implicit assumptions on the reliability, trustworthiness and scale of participation. As discussed in the introduction, subversion of these assumptions can be catastrophic. A lost guest output or gradient is a single point of failure. The intersection $\mathcal S_{\mathrm{labels}}\cap \mathcal S_{\mathrm{guests}}$ may not be meaningfully large, especially when there are many guests. And, it is possible for a malicious guest to infer private data from feedback. We further analyze these shortcomings here.

\subsubsection{Single Points of Failure}
\label{spof}
In order to make a prediction on a batch of data, the host model $m_h: \mathbb R^{k_1}\times\dots\times\mathbb R^{ k_{|\mathcal{G}|}}\to \mathbb R^{\text{out}}$ requires inputs from all guest models. If an input is unavailable to the host model, no prediction can be made-- a catastrophic fault. Similarly, local models' updates require gradients from the host, lacking which the model convergence slows down and may fail altogether. We consider the following fault model for a round of training:
\begin{itemize}
    \item \textit{Guest faults}: A guest may be unable to compute its output, update its model, or communicate.
    \item \textit{Host faults}: The host itself may fail, and be unable to receive inputs or calculate gradients.
    \item \textit{Connection faults}: The communication link between a guest and the host may be dropped during the forward or backward pass.
\end{itemize}
From a VFL host's perspective, a guest fault and connection fault manifest as a missing input to its model $m_h(\text{ }\cdot\text{ }; \theta_h)$ during training. The host may employ one of two strategies:

\begin{enumerate}
\vspace{-2ex}
    \item \textit{Wait}: The host may wait until all input arrives.
    \begin{itemize}[leftmargin=*]
        \item Under a guest fault or connection fault, the host will remain waiting indefinitely, and training will not continue.
        \item If the host polls an unresponsive guest, it may receive an input intended for the next iteration, which will lead to incorrect model evaluation.
    \end{itemize}
\vspace{-2ex}
    \item \textit{Timeout}: The host may set a deadline for each round.
    \begin{itemize}[leftmargin=*]
        \item Under a guest fault or connection fault, training will fail since the host model is missing input.
        \item If a message is successful but late, the model is still missing input.
        \item If the model skips that training round and moves onto the next, it may receive the late message from the previous round, which will lead to incorrect model evaluation.
    \end{itemize} 
\end{enumerate}

\vspace*{-.15in}
Therefore, a connection or guest fault leads to a catastrophic failure for training.
 When the host fails, the guest models are unable to update their parameters. Since the cardinality of the joint training dataset is that of the intersection of participants' sample spaces $|\mathcal{S}_{\mathrm{guests}}\cap \mathcal{S}_\mathrm{labels}|$ (which may be small), losing updates from a single batch of data may be detrimental to model performance.
\subsubsection{Data Intersection}
Since VFL training is contingent on entity alignment, it is not usable in scenarios where the intersection of entities known to each guest is small. This is often the case when the number of guests is large. It could also be the case that, although $\mathcal{S}_{\mathrm{guests}}$ is reasonably large, labels are not available for all its members, i.e., $\mathcal{S}_\mathrm{labels}\cap \mathcal{S}_{\mathrm{guests}} \subset \mathcal{S}_{\mathrm{guests}}$. 
\subsubsection{Inference Attacks}
\label{sec:Inference}
\citet{LabelInf} demonstrate that it is possible for a malicious agent to directly 
infer the true class from gradients (e.g. by analyzing their sign) for many common 
loss functions during training. If a curious guest inflates its learning rate, the 
joint model gives preference to its outputs. If it can obtain even a small number of labels, 
it may train its own prediction head. The authors show that such a malicious model can make meaningful guesses for labels. 

\citet{Luo2021, Erdoan2022} show that a generative network can be trained by a set of 
guests to model the distribution of other guests' features. The authors also show that a
malicious guest only needs to make a good guess for labels to directly infer features of 
other guests in vertically federated linear classifiers and decision trees.

Typical defenses include differential privacy, gradient compression, homomorphic encryption
and secret shares for gradients and model features. However, these methods usually have a
negative effect on model convergence rate and tightness \citep{khan2023vertical}. 
Moreover, implementing such cryptographic algorithms is computationally expensive, and scales poorly.

\section{Related Work}
In this section, we review some closely-related lines of research which have motivated DVFL. A more verbose discussion is available in the Appendix
\ref{appendix:rw}.
\label{sec:related}
\paragraph{Fault Tolerant Distributed Optimization} Like any other distributed system, redundancy
enables fault resilience and tolerance in distributed optimization \citep{Liu2021}.
\citet{liu2021survey,bouhata2022byzantine} provide extensive reviews of Byzantine fault tolerant
horizontally distributed learning algorithms. Most methods hurt model performance and carry
significant computational overhead. 

\textbf{For VFL:} For VFL, there is limited work on crash faults. However, there is a line of work on the related case of straggling guests. That problem definition only extends to the count of failures being bounded as in FedVS \citep{FedVS}, and/or late messages eventually arriving \citep{CVFL, Shi2022, Zhang2021SecureBA, Hu2019}. A real system could have an arbitrary number and duration of crashes, so these methods do not generalize to crash faults. Typically, they do not address faults in the backward pass. Paradigms that require periodic synchronization e.g. Gradient Assisted Learning \citep{diao2022gal} are similarly hamstrung. 

\paragraph{Asynchronous VFL} There are some efforts in speeding up training by relaxing synchronization in and splitting effort of gradient calculation. But many of these methods expect all inputs during the forward pass \citep{Wang2024, Shi_2024, Zhang2021SecureBA, Gu2021PrivacyPreservingAV}. \citet{chen2020vafl} present a VFL algorithm which has an asynchronous forward pass, which is a viable candidate for crash tolerance. Another line of work \citep{Zhang2021SecureBA, Gu2021PrivacyPreservingAV} has an implicitly asynchronous forward pass-- its aggregation strategy is summation rather than concatenation.

\paragraph{Federated Transfer Learning} Federated Transfer Learning (FTL) is a strategy for systems 
with overlap in both sample space and feature space.  In \citet{Liu2020, feng2020multiparticipant,10.1016/j.knosys.2022.109384}, 
parties learn a common latent space using local feature extractors and transfer to their tasks. 
For vertically partitioned datasets with small $|\mathcal{S}_\mathrm{labels}|$ but 
large $|\mathcal S_\mathrm{guests}|$, \citet{Sun_2023_ICCV} introduce Few-Shot VFL. 
A few epochs of training on the aligned dataset are followed by clustering-based 
semi-supervised learning on the unlabelled entities of each client's dataset. Besides 
the unwanted computational overhead required to generate the labels, this is effectively a label inference attack.

\paragraph{Greedy and Localized Learning} Recently, there has been a resurgence of 
interest in algorithms which train modules of NN layers on local objectives as opposed
to BP. One reason is inherent locking in BP, described in \citet{pmlr-v70-jaderberg17a}. 
With local supervision, a layer need not wait for the global backward signal to update 
its model (``backward unlocking"). \citet{pmlr-v119-belilovsky20a} propose a replay buffer: 
Each module of layers stores its inputs in a replay buffer and forward passes its front. 
It need not wait for a fresh forward signal to compute an output (``forward unlocking"). 
Note that \emph{forward and backward locking at the split layer are exactly the failure points for VFL}. 
Local updates have proved useful for practical HFL (e.g. \citet{collins2022fedavg, 9952869, briggs2020federated}), 
where labels are available to guests. For VFL, \citet{Fu2022} use cached statistics to construct local gradients. 
As far as we know, DVFL is the first with fully localized model training.

\section{Why Not an Engineering-Level Solution?}
\label{in-VFL-mitigation}
Keen observers will note that the critical failures of the ``\textit{wait}" and ``\textit{timeout}" strategies discussed in \S\ref{spof} can be partially mitigated with some extra bookkeeping. If the guest sends back the sample index (or indices of samples in its batch) and epoch number with each message, the host can uniquely identify which request it belongs to. Now \textit{timeout} may be employed with a strategy to fill missing inputs; late messages may be identified and discarded. \textit{Wait} may be employed by periodically repeating unfulfilled requests; messages from other iterations may be queued. So why DVFL?

Notably, neither strategy handles faults on on backward passes. Moreover, \emph{wait} is a busy-wait. Besides being incompatible with deadline-based scheduling and wasting computational and communication bandwidth with repeated unserviced requests, there is no guarantee that training ever continues. In fact, this would also be the case for event-driven computation. But it is not so for DVFL. As for \textit{timeout}, it is not trivial to choose a good replacement for missing inputs. We investigate some obvious options and show that DVFL outperforms them. 

\section{Proposed Method}
\label{sec:prop}
DVFL splits the training into two phases coordinated by the owner. The first is an unsupervised and decoupled greedy \emph{guest-host} representation learning phase. In this phase, the guest and host models learn feature extractors on their own unsupervised objectives. The second is a supervised \emph{owner} transfer-learning phase where the label owner learns to aggregate encodings from the host space and produce label predictions. We do not consider faults in owner training. They do not affect the guests and hosts which train the bulk of the network. Even if the owner were unstable, it only needs to collect hosts' final encodings for each input once-- then, it may train its model offline whenever it is up.\footnote{This raises the question-- why not do this between the guest and the host as it is in FedOnce \citep{Wu2022}? Firstly, the owner's job is only to predict, so a deep model is not required. As such, the dimensions of the embeddings and the computational cost of training its model will typically be small enough to neglect the cost of retrial, unlike a FedOnce/VFL host. Moreover, if the aggregator only gets one round of communication as in FedOnce, model performance is hindered. We explore this in our experiments and the Appendix.}

Since guests do not rely on feedback from hosts to update their models, host or connection faults during the backward pass are irrelevant to the guest models. Hosts store activations from guests as they arrive and use the latest message for their forward pass, so a missing input on the forward pass is no longer critical.

The lack of end-to-end feedback combined with the three-tier hierarchy leads to significant robustness. The task of a host's model is only to learn an encoding of features in the joint guest domain, which is not an instance-specific task. Therefore, a host may still update its model meaningfully with some out-of-date activations. Similarly, guests may use their entire datasets as opposed to only members of the intersection. Besides leading to better models, using more data means a single fault does not affect the overall uptime as much. On the backward pass, faults only affect the crashed participant.


\subsection{The DVFL Hierarchy}
\textbf{Guests:} The role of a guest in DVFL is identical to a VFL guest's (see \S\ref{sec:hierarchy}), i.e. guests $g_i\in \mathcal G$ are the agents $g_i$ who own datasets $\mathcal D_i$ with records of features $\mathcal F_i$ for the set of entities whose indices are present in $\mathcal{S}_i$. The guest model's role is still to {encode its dataset's private features}. Unlike VFL, they learn to do so on local unsupervised objectives. 

\textbf{Hosts:}
Hosts are members of the set $\mathcal H$ of agents $h_i$ which own aggregating \emph{host} models, which accept the concatenation of all guest model outputs as input. Each $h_i\in \mathcal H$ has: 
\begin{enumerate}
\vspace{-2ex}
    \item Host model $m_{h,i}(\text{ }\cdot\text{ }; \theta_{h,i}):\mathbb R^{k_1}\times\mathbb R^{k_2} \times \dots\times\mathbb R^{ k_{|\mathcal{G}|}}\to \mathbb R^{k_{h, i}}$ whose task is to aggregate and compress the data it receives from guests. It has its own unsupervised objective.
    \vspace{-1ex}
    \item Single-reader single-writer ``communication" registers $\mathbf B_{j, i}\forall j\in\{1, 2, \dots, |\mathcal{G}|\}$. Each guest has a register on $h_i$, and writes its output to it. The host reads it to form its model input.
    \vspace{-1ex}
    \item Input replay buffer $\mathbf A_{i}$. The host $h_i$ uses it to store a history of model inputs if it wishes to reuse them.
    \vspace{-1ex}
\end{enumerate}
Each host model has the same architecture but its own independent parameters $\theta_{h,i}$. Unlike standard VFL, the hosts do not make predictions. Their models are only responsible for aggregating guest outputs and learning to effectively map them to a lower dimensional space. Note that when there is no such compression, the host can achieve this without any learnable parameters-- only a single aggregating layer e.g. concatenation is necessary.

\textbf{Owner:} The owner agent $o$ executes each training and communication round. It has access to ground truth labels, and is responsible for training a ``final" or ``owner" model $m_o(\text{ }\cdot\text{ }; \theta_o): \mathbb R^{k_{h, 1}}\times\dots R^{k_{h, |\mathcal H|}}\to\mathbb R^{\text{out}}$ on 
 representations from members of $\mathcal H$ (i.e. it is a transfer learning model). The owner has access to the ground truth targets $L = \left\{\mathbf y_j \in \mathbb R^{\text{out}} : \text{label }\mathbf y_j\text{ of } \mathbf x_j, j \in \mathcal{S}_\mathrm{labels}\right\}$.

 \begin{figure}
    \centering
    \includegraphics[height=0.15\textheight]{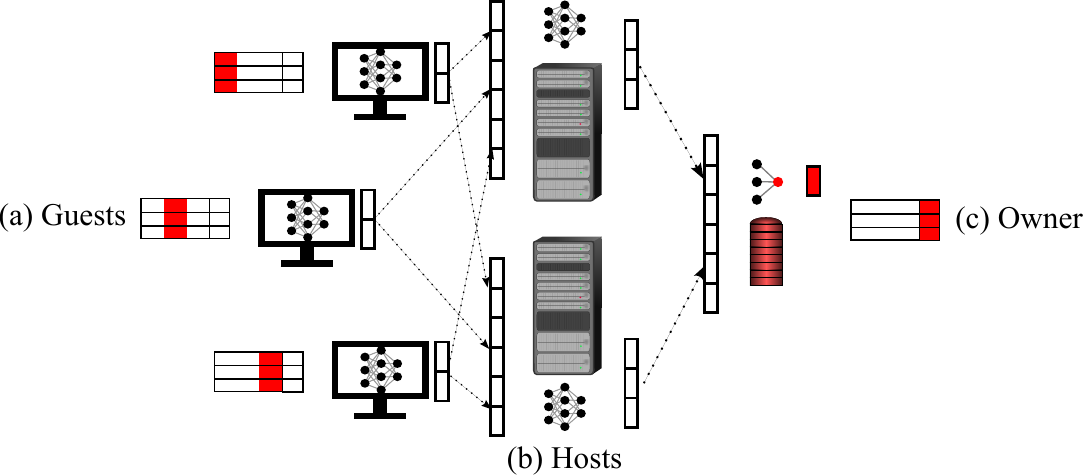}
    \caption{A DVFL system with three guests (a) and two hosts (b). Guests train their local models on unsupervised objectives and hosts also train their aggregating models on unsupervised objectives. After \if those models are trained\fi that, the label owner (c) trains a transfer learning model (such as a linear classifier head) on the encodings from the hosts.}
    \label{fig:dvfldiagram}
    \vspace{-3ex}
\end{figure}
\subsection{Asynchronicity in DVFL}
To avoid a single point of failure, host models should continue training without inputs from some guests. Single-reader single-writer ``communication" registers allow hosts to reuse old activations under guest and communication faults. Hosts can store model input history in replay buffers for reuse if they wish to train for more iterations than they receive activations for. We discuss these mechanisms in further detail in Appendix \ref{appendix:complexity}.

The task of a host's model is only to learn an encoding of features in the joint guest domain, which is not an instance-specific task. Therefore, a host may still update its model meaningfully with some out-of-date activations.
\subsection{Data Beyond the Intersection}
\if Until now, we have assumed the input datasets to each model are ``aligned", implying they only contain records for entities whose indices are in $\left(\bigcap_{i=1}^{|\mathcal{G}|}\mathcal{S}_\mathrm{labels}\right)\cap \mathcal{S}_\mathrm{labels}$. Now, we present a strategy for owner training on categorical tasks which will allow each dataset to be of size $\displaystyle\min_{i\in\{1,\dots, |\mathcal{G}|\}}|\mathcal{S}_\mathrm{labels}\cap \mathcal{S}_\mathrm{labels}|$. 

We extend that principle to the task-agnostic unsupervised learning of host and guest models and show that guests can use their entire datasets regardless of the final owner's labels.

\textbf{Cutmix}: Cutmix is a data augmentation commonly used in categorical computer vision tasks. Suppose the task is binary classification between dogs (say one-hot label $[0,1]$) and cats (say label $[1,0]$), and our training dataset contains some labeled $n\times n$ images of dogs and cats. A $p\times q$ patch of a dog image is cut and pasted on top of a cat image, and replaced with a $p\times q$ patch of a cat image. This is effectively a replacement of $pq$ of the $n^2$ features with features from an instance of the other label. The label of the modified dog image is interpolated to $[\frac {pq}{n^2}, 1-\frac{pq}{n^2}]$ and that of the modified cat image to $[1-\frac{pq}{n^2}, \frac {pq}{n^2}]$. 

Not only does an NN which successfully trains on the raw dataset also train on the augmented dataset, but the model trained on the augmented dataset performs much better on test datasets. Intuitively, the hidden layers of the NN learn to model a more general distribution, one which can generate dog-cat hybrid samples. Similar label-mixing data augmentation techniques exist for non-categorical tasks, such as RegMix for regression.

\textbf{Entity Augmentation}: To train the owner model, we may similarly construct samples of artificial entities by concatenating features of different entities and averaging their labels\footnote{With reference to Cutmix, an unweighted averaging of labels assumes that the sets of features belonging to all guests have equal contribution to the final aggregation. This means the initial layers of the owner model must also learn to correctly weight each feature from the host latent space.}. Algorithm \ref{alg:DataAug} describes a naive technique to achieve such ``entity augmentation" for categorical tasks, which allows the owner model to be trained on $\displaystyle\min_{i\in\{1,\dots,|\mathcal{G}|\}}\left|\mathcal{S}_\mathrm{labels}\cap \mathcal{S}_\mathrm{labels}\right|$ samples. Once the label owner has decided the order of entities for each guest, it can inform them. The effect is a dataset $\mathcal D_i''$ which is both ``entity aligned" and ``entity augmented" with respect to the other guests and the label owner.

\begin{algorithm}[tb]
   \caption{Entity Alignment \& Augmentation for categorical tasks}
   \label{alg:DataAug}
   \begin{algorithmic}
      \STATE {\bfseries Input:} $\mathcal{S}_\mathrm{labels}\forall i\in\{1,2,\dots,|\mathcal{G}|\}$  \COMMENT{Set of entities for which guest $i$ has features}, $\mathcal{S}_\mathrm{labels}$  \COMMENT{Set of entities for which label owner has labels}
      \STATE Label set $L = \left\{\mathbf y_k \in \mathbb R^{c}\;\middle|\; \mathbf y_k \text{ = one-hot label for entity $k$}\right\}$
      \STATE Initialize empty queues for each guest $Q_i$ and an empty label queue $Q_{\text{label}}$
      \FOR{$i$ from 1 to $(|\mathcal{G}|)$ in parallel}
        \STATE $\mathcal{S}_{i,I} \gets \text{PSI}(\mathcal{S}_\mathrm{labels}, \mathcal{S}_\mathrm{labels})$  \COMMENT{Private Set Intersection}
      \ENDFOR
      \STATE $\mathcal{S}_{\text{alignable}} \gets \mathcal{S}_{1,I}$  \COMMENT{Initialize with the first guest's intersection}
      \FOR{$i$ from 2 to $|\mathcal{G}|$ in parallel}
        \STATE $\mathcal{S}_{\text{alignable}} \gets \text{PSI}(\mathcal{S}_{\text{alignable}}\cap \mathcal{S}_{i,I})$
      \ENDFOR
      \STATE{// Entity Alignment}
      \FOR{each entity $k$ in $\mathcal{S}_{\text{alignable}}$}
        \FOR{$i$ from $1$ to $|\mathcal{G}|$ in parallel}
          \STATE Remove $k$ from $\mathcal{S}_{i,I}$ and add it to $Q_i$
        \ENDFOR
      \STATE Add the label of $k$ to $Q_{\text{label}}$
      \ENDFOR
      \STATE{// Entity Augmentation}
      \WHILE{$\mathcal{S}_{i,I} \neq \phi\text{ }\forall i$}
      \STATE label $\gets 0$
        \FOR{$i$ from 1 to $|\mathcal{G}|$}
          \STATE Remove $k_i$ from $\mathcal{S}_{i,I}$ and add it to $Q_i$
          \STATE label $\gets$ label$+\mathbf y_k$
        \ENDFOR
        \STATE label $\gets \frac{\text{label}}{|\mathcal{G}|}$
        \STATE Add label to $Q_{\text{label}}$
      \ENDWHILE
   \end{algorithmic}
\end{algorithm}

Regardless of whether the labels pertain to a categorical task, we may use this principle for the unsupervised training of guest and host models. We posit that even if guests provide hosts with latent encodings pertaining to differing entities, the host model learns a usable feature extractor. In Section , we demonstrate this by intentionally misaligning samples before training.

Effectively, each guest dataset can be partitioned into three. Assume for simplicity that $|\mathcal{S}_\mathrm{labels}| = |S| \text{ (say) }\forall i \in\{1, 2, \dots, |\mathcal{G}|\}$. Within each dataset $\mathcal D_i$, there are $|\mathcal{S}_{\mathrm{guests}}\cap \mathcal{S}_\mathrm{labels}| = |\mathcal{S}_\text{alignable}|\text{ (say)}$ ``alignable" samples, $\displaystyle\min_{i\in\{1,\dots,|\mathcal{G}|\}}\left|\mathcal{S}_\mathrm{labels}\cap \mathcal{S}_\mathrm{labels}\right| = |\mathcal{S}_{\text{aug}}|$ (say) ``augmentable" samples. The remaining $|S|-|\mathcal{S}_{\text{aug}}|-|\mathcal{S}_{\mathrm{guests}}\cap \mathcal{S}_\mathrm{labels}|$ are ``unalignable" samples. 

After entity alignment and augmentation, the owner has constructed an queue of entities $Q_i$ for each guest $g_i$. It may inform each guest of its $Q_i$ before or during owner training. 
\fi

Until now, we have assumed the input datasets to each model are ``aligned", viz. they only contain records for members of $\mathcal{S}_\mathrm{guests}\cap \mathcal{S}_\mathrm{labels}$. The learning target of the guest models are to compress any data they receive as input, i.e. to encode features of a realization of guest domain's data distribution. Similarly, each host learns to aggregate and compress encodings from all guests. These tasks depend on the input data's domain (i.e. not on its instance). Therefore, guests may train their local models on features of $\mathbf x_j$ even when $j\in \mathcal{S}_{\mathrm{guests}}\setminus \mathcal{S}_\mathrm{labels}$.

\subsection{DVFL for Split Training of Neural Networks} NNs are natural choices for participants' models. Unsupervised neural networks are feature extractors \citep{Intrator1992,Becker1996}, which is precisely the role of guest and host models. Similarly, the strong task performance of NNs have been the primary driver of deep learning over the past few years, so they are suitable for owners, too. Indeed, just as SplitNN is a popular implementation of VFL, we expect NNs to be common models for DVFL participants. 

Figure \ref{fig:SplitNN} illustrates the relationship between DVFL and VFL in the context of a global NN model. We may say $m_i$s are all NNs with parameters $\theta_i$, $m_{h,j}$ are also NNs whose input is the concatenation of the outputs of all $m_i$s, and $m_o$ is an NN whose input is the concatenation of the outputs of all $m_{h,j}$s. The pseudocode is presented in the Appendix \ref{appendix:dvflnn} (Algorithm \ref{alg:DVFL}).
\section{Experiments}
\label{sec:expts}
\paragraph{Datasets}
As a toy example, we consider a system of $|\mathcal{G}|$ guests which can each see $\frac1{|\mathcal{G}|}^{th}$ of an handwritten digit from the MNIST dataset \citep{deng2012mnist}. For our experiments, we set $|\mathcal{G}| = 4$ i.e. $g_1$'s dataset contains the top $28\times 7$ pixel patch of an MNIST image, $g_2$'s dataset contains the second \if $28\times 7$ patch\fi, $g_3$'s contains the third, and $g_4$'s contains the bottom $28\times 7$ patch (see \if Figure\fi Fig. \ref{fig:MNIST}). We also experiment on some benchmark VFL datasets in the Appendix.
\paragraph{New Baselines and Loss Functions} 
DVFL is an alternate strategy to VFL. Naturally we compare DVFL for NNs to SplitNN, a direct application of VFL to NNs. Naive VFL implicitly employs the “wait” protocol discussed in \S\ref{spof}.

We also try the modified timeout strategy from \S\ref{in-VFL-mitigation} by imputing missing inputs with zeros (zeros), stale activations (buffer), and skipping that round altogether (skip). For SplitNN hosts and DVFL owners, we use cross-entropy loss. For DVFL hosts and guests, We use a learnable decoder head and MSE reconstruction as an unsupervised objective. 

\paragraph{FedOnce Baseline} We also provide a baseline with FedOnce \citep{Wu2022}. FedOnce is similar to DVFL in the sense that guests train their own models on unsupervised objectives, and similar to VFL in the sense that there is a centralized aggregating host that owns labels. FedOnce only requires one round of communication (the guest-encoded dataset after all guests have fully trained). It can be argued that each guest can retry until the host has heard from all of them (since it would not be holding up training of any future iterations). If that is implemented, faults are not a worry. Regardless, we also try an alternate fault-handling strategy inspired by VFL-zeros for compatibility with deadline-based scheduling.


Hyperparameters, model details and ablation study are presented in the Appendix.
\paragraph{Asynchronous Baselines} We also compare DVFL to the two discussed viable asynchronous VFL algorithms: VAFL \citep{chen2020vafl} and VF\textbf{B}$^2$ \citep{Zhang2021SecureBA}. For VAFL, we eschew random batch sampling for iteration (in the same order by all guests). This locality gives us a skyline of VAFL's performance-- non-active guests' ``stale" activations will be recent. For VF\textbf{B}$^2$, we simulate without the two distinct communication graphs to minimize the scope for faults. Further, since we experiment with only one label-owning party, the ``active parties" class only contains a single host ($m=k=1$ by the convention of the original paper). Also, since the aggregation strategy is summation, we modify the first layer of the host model's input dimension to equal a single host's output dimension. 
\paragraph{FedVS Baseline and More Datasets}
We also compare the model performance of SplitNN and DVFL to results cited in the work FedVS \citep{FedVS}. Since FedVS is incompatible with our fault model, we only compare performance under perfect conditions. We use the same datasets (\citet{YEH20092473,uci,li_andreeto_ranzato_perona_2022,Sakar2019ACA}) as used in the experiments for FedVS. FedVS restricts its models to use polynomial networks \citep{livni2014algorithm} i.e. equally-sized linear layers with polynomial activation functions. The models we use for SplitNN and DVFL are smaller-- each layer is half as wide as the next, and we use Leaky ReLUs. 
\subsection{Fault Simulation}
Six hyperparameters control the fault model described in \S\ref{spof}. $R^{(d)}_i$ is the fault rate for a live 
$i\in \{\mathrm{connection},\mathrm{guest}, \mathrm{host}\}$. $R^{(u)}_i$ is the chance a dead $i$ rejoins the network before the next request.
We discuss how exactly the faults are simulated in the Appendix. We assume no faults during label supervision or inference.
\if
We find that a guest latent dimension of 80 encodes sufficient semantic information for classification.
\subsubsection{Guest Models}
The guest models are linear autoencoders with hidden layers of dimension $\frac{400}{|\mathcal{G}|} = 100$, $\frac{320}{|\mathcal{G}|} = 80$ and  $\frac{400}{|\mathcal{G}|} = 100$ respectively. We use a ReLU nonlinearity at the middle layer, which we use as the encoded guest output. A sigmoid nonlinearity is used to constrain the reconstruction to $[0, 1]$.
\subsubsection{Host Models}
The host models are also linear autoencoders, with hidden layers of dimension $200$, $160$ and $200$ respectively. We use a Leaky ReLU (with slope $-0.01$ in the negative half-plane) nonlinearity at the middle layer, whose output we use as the encoded host output. A ReLU nonlinearity is used to constrain the reconstruction to $[0,\infty)$. 
\subsubsection{Label Owner Model}
The label owner has a $10-$class linear classification head with hidden layers of dimension $160$ and $40$ with a Leaky ReLU nonlinearity at the second hidden layer.
\subsubsection{SplitNN Baseline}
The encoders and classification head together effectively form a classifier multi-layer perceptron with hidden layers of dimension $400, 320, 200, 160, 40$. We also train a SplitNN baseline with this global architecture, i.e. guest models are identical to the DVFL encoders, and the host model has hidden layers of dimension $200, 160, 40$.
\subsubsection{Hyperparameters}
Unless it is otherwise mentioned, we train guest models for 20 epochs, host models for 40 epochs and owner models for 60 epochs. We train the SplitNN baseline for 60 epochs. For the guest, host, and SplitNN models, we use an Adam optimizer with learning rate $10^{-3}$ and a weight decay factor of $10^{-6}$.
\fi
\subsection{Experimental Setup}
\begin{table*}[]
\caption{We compare DVFL to standard VFL. After filtering outliers by IQR, we report the mean and 2x the standard deviation of MNIST test performance when trained with various entities $i\in\{\mathrm{connection, guest, host}\}$ susceptible to fail at a rate of $R^{(d)}_i = 0.3$. A failed resource has $R^{(u)}_i = \mathbb P_{\mathrm{rejoin}} \in \{1, 0.5, 0.1\}$ to be available next time it is requested. SplitNN, the baseline VFL algorithm, cannot train at all when there are any faults. On the other hand, our DVFL implementation degrades gracefully with faults. Using an explicit fault handling strategy within VFL does allow it to train. But the degradation with faults is sharper than DVFL and the variance in model performance is higher. Under fault-free circumstances, SplitNN slightly outperforms our DVFL implementation. With minor adjustments, SplitNN can tolerate some faults, but our approach is more resilient and stable.}
\label{tab:my-table}
\resizebox{\textwidth}{!}{%
\begin{tabular}{@{}ccccccccccc@{}}
\toprule
\multirow{4}{*}{\textsc{\textbf{Strategy}}} & \multicolumn{10}{c}{\textsc{\textbf{Test Accuracy (\%) with 0.3 Loss Rate}}}                                                                                                    \\ \cmidrule(l){2-11} 
                          & \textsc{No Faults}          & \multicolumn{3}{c}{\textsc{Connection Loss}}                      & \multicolumn{3}{c}{\textsc{Guest Loss}}                     & \multicolumn{3}{c}{\textsc{Host Loss}}                      \\ \cmidrule(lr){2-2}\cmidrule(lr){3-5}\cmidrule(lr){6-8}\cmidrule(l){9-11}    
                          & $\mathbb P_{\mathrm{rejoin}} =$ N/A  & $\mathbb P_{\mathrm{rejoin}} =$ 1    & $\mathbb P_{\mathrm{rejoin}} =$ 0.5  & $\mathbb P_{\mathrm{rejoin}} =$ 0.1  & $\mathbb P_{\mathrm{rejoin}} =$ 1    & $\mathbb P_{\mathrm{rejoin}} =$ 0.5  & $\mathbb P_{\mathrm{rejoin}} =$ 0.1  & $\mathbb P_{\mathrm{rejoin}} =$ 1    & $\mathbb P_{\mathrm{rejoin}} =$ 0.5  & $\mathbb P_{\mathrm{rejoin}} =$ 0.1  \\ \toprule
DVFL-NN & 97.80$\pm$\textbf{0.12} & \textbf{97.78}$\pm$\textbf{0.11} & \textbf{97.78}$\pm$\textbf{0.14}      &\textbf{97.72}$\pm$\textbf{0.17}      & \textbf{97.77}$\pm$\textbf{0.11} & \textbf{97.74}$\pm$\textbf{0.16}      & \textbf{97.58}$\pm$\textbf{0.20}      & \textbf{97.81}$\pm$\textbf{0.12} & \textbf{97.79}$\pm$\textbf{0.14}      & \textbf{97.60}$\pm$\textbf{0.33}      \\
VFL (SplitNN)             & 97.85$\pm$0.17 & N/A           & N/A           & N/A           & N/A           & N/A           & N/A           & N/A           & N/A           & N/A           \\
VFL (SplitNN-skip)        & \textbf{97.86}$\pm$0.16 & 97.50$\pm$0.54 & 96.60$\pm$0.51 & 77.21$\pm$5.34 & 97.44$\pm$0.57 & 96.65$\pm$0.45 & 80.22$\pm$4.31 & 97.65$\pm$0.41 & 97.59$\pm$0.50 & 97.29$\pm$0.89 \\
VFL (SplitNN-zeros)       & 97.84$\pm$0.15 & 97.64$\pm$0.38 & 97.53$\pm$0.40 & 97.44$\pm$0.36 & 97.68$\pm$0.40&97.59$\pm$0.40 & 96.95$\pm$0.50  & 97.68$\pm$0.30 & 97.65$\pm$0.37 & 97.31$\pm$0.93 
\\
VFL (SplitNN-buffer)      & 97.85$\pm$0.17      & 97.66$\pm$0.38      & 96.85$\pm$0.74      & 93.45$\pm2.10$ & 97.59$\pm$0.58     & 96.76$\pm$0.70      & 91.78$\pm$1.70      & \textbf{97.81}$\pm$0.32      & 97.71$\pm$0.40      & 97.69$\pm$0.51
\\
VFL (FedOnce-zeros) & 97.76$\pm$0.20 & 96.89$\pm$0.29 & 96.41$\pm$0.30 & 92.80$\pm$0.75 & 97.47$\pm$0.37 & 97.45$\pm$0.31 & 97.50$\pm$0.31 & 97.28$\pm$0.35 & 97.34$\pm$0.37 & 96.93$\pm$0.51
\\
VAFL (skyline)    & 97.57$\pm$0.30 & 97.17$\pm$0.46 & 96.88$\pm$0.36 & 96.46$\pm$0.44 & 97.18$\pm$0.42 & 96.99$\pm$0.44 & 96.45$\pm$0.51  &97.17$\pm$0.48 &96.89$\pm$0.36 &96.42$\pm$0.52
\\ 
VF\textbf{B}$^2$ (skyline) & 97.72$\pm$0.16 & 97.43$\pm$0.38 & 97.33$\pm$0.41&96.09$\pm$0.47&97.36$\pm$0.45&97.24$\pm$0.41&96.29$\pm$0.42&97.43$\pm$0.49&97.20$\pm$0.61&96.79$\pm$0.90\\\bottomrule
\end{tabular}%
}
\end{table*}
\paragraph{Performance under Faults}
In order to show that it is possible to use DVFL even with crash faults, we train DVFL with $|\mathcal{H}| = 4$ hosts with a fault rate of 0.3 and a rejoin rate of 0.1. For a given experiment, the failure mode (connection, guest, or host) is fixed. 

$R^{(u)} = 1$ simulates single-message loss, since the next time a message is sent or received, the participant is sure to have rejoined. $R^{(u)} = 0.5$ simulates short bursts of loss. $R^{(u)} = 0.1$ simulates appreciable downtime. 

\paragraph{Effect of Redundancy}
In order to show that redundancy of hosts improves model performance and fault tolerance, we  simulate DVFL training with $|\mathcal{H}|$ = 1, 2, 3, and 4 hosts. We measure test accuracy given connection fault rate $\in \{ 0, 0.3, 0.6\}$ during training.

\paragraph{Learning from a Limited Intersection}
We first fix a small labelled and aligned intersection size. The remaining sample space is shuffled and distributed evenly across all four guests. For example, if we wish to use $1024$ labeled samples, we shuffle samples indexed $1024-59999$ and split the $58976$ samples evenly, such that each guest gets $14744$ unique MNIST patches + $1024$ intersecting MNIST patches = $15768$ total samples. We try this experiment with $128$, $256$, $512$ and $1024$ labeled samples. We compare with SplitNN trained on only the labeled samples.
\section{Results and Discussion}
\begin{figure*}[]
    \centering
    \subfigure[For our experiments, the input received by each guest for a given sample, as well as the reconstruction its autoencoder model produces.]{
        \resizebox{!}{0.165\textheight}{
        \includegraphics[width=0.1\columnwidth]{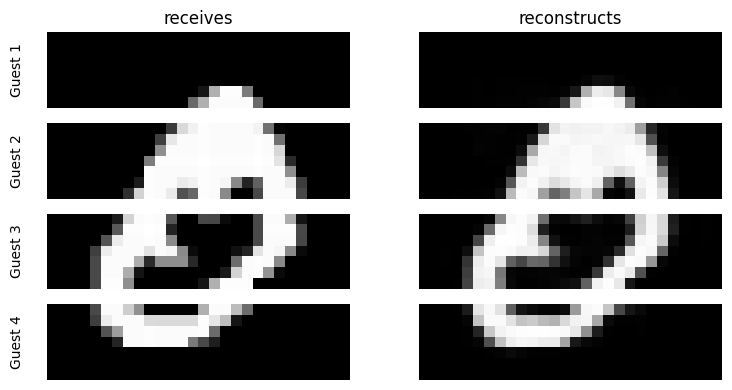}
        \label{fig:MNIST}}
    }
    \hspace{1ex}
    \subfigure[Model performance improves with redundancy, especially when there is a nonzero crash rate.]{
        \resizebox{!}{0.23\textheight}{\begin{tikzpicture}
        \begin{axis}[
            axis lines = left,
            xlabel={Number of Hosts},
            ylabel={MNIST Test Accuracy (\%)},
            xmin=0, xmax=5,
            ymin=97.00, ymax=98.00,
            xtick={0,1,2,3,4,5},
            ytick={97.30, 97.40,97.50,97.60,97.70,97.80,97.90},
            legend pos=south west,
            ymajorgrids=true,
            grid style=dashed,
        ]
        \addplot[
            color=blue,
            mark=square,
            ]
            coordinates {
            (1,97.74)(2,97.79)(3,97.80)(4,97.82)
            };
            \legend{\(R_{\mathrm{connection}}\) = 0}
        \addplot[
            color=red,
            mark=asterisk,
            ]
            coordinates {
            (1,97.60)(2,97.70)(3,97.74)(4,97.79)
            };
            \addlegendentry{\(R_{\mathrm{connection}}\) = 0.3}
        \addplot[
            color=orange,
            mark=diamond,
            ]
            coordinates {
            (1,97.52)(2,97.65)(3,97.70)(4,97.74)
            };
            \addlegendentry{\(R_{\mathrm{connection}}\) = 0.6}
        \end{axis}
        \end{tikzpicture}
        \label{fig:faults}}
    }
\end{figure*}

\subsection{Graceful Degradation with Faults}
(Table \ref{tab:my-table}) shows that model performance degrades with faults, but not catastrophically (unlike VFL with SplitNN). In ideal conditions, DVFL performance is slightly poorer than SplitN's. This is not unexpected as we will discuss in \S\ref{sec:lim} Observe that the degradation with faults in DVFL is much more graceful and \emph{stable} than the SplitNN variants.

In the case of SplitNN-based algorithms, a connection and guest fault are functionally identical. In the case of DVFL, there are some performance savings under connection faults-- the corresponding guest and host can still meaningfully learn even in the absence of connection. Further, redundant hosts may continue training as normal.

\paragraph{Tolerating some faults pays off}
The artifical input induces a Dropout effect \citep{JMLR:v15:srivastava14a}. This reduces bias but competes with information loss in the model. DVFL models, without cross-device feedback, see smaller information loss compared to SplitNN-based methods (which do not respond to faults as well, but still exhibit somewhat graceful test accuracy degradation). The simplest SplitNN-based method, \emph{Skip}, leads to the worst performance since information from non-faulty guests is also lost. \emph{Zeros} works as if failed neurons do not exist at all in simple linear and convolutional layers. This would fail with more complex architectures and algorithms e.g. skip connections and constrained solvers. \emph{Buffer} empirically cannot exploit locality in activations and performs poorly. Reconstructing missing inputs based on statistics, as \citet{Fu2022} do for gradients, may be the best solution but would suffer from bias and computational infeasibility.

\paragraph{Comparison to asynchronous VFL} Asynchronous algorithms suffer from higher variance, perhaps since the information loss from a single failure do not have the safety net of information from successful interactions. Further, these skylines perform significantly worse than VFL and DVFL even when there are no faults. In fact, we argue that VFL-zeros provides a skyline for performance from any asynchronous algorithm (at least when the model has no skip connections or optimization constraints). As discussed in the previous paragraph, VFL-zeros acts as if failed neurons did not exist at all. This is as if an oracle told an asynchronous host which guests/connections will be live, and the host calculated losses based on multiple guests' embeddings at once hence leading to lower-variance loss estimates. VF\textbf{B}$^2$ and other algorithms using summation-based aggregation also suffer from constraints on the model architecture. The output of each guest must have the same dimension. Further, the host input dimension is smaller than what it would be if it were to aggregate by concatenation, so there are fewer degrees of freedom for the model.

\paragraph{Comparison to FedVS and results on other datasets} FedVS \citep{FedVS} provides both gradient privacy (via cryptographic multi-party communication) and straggler resistance. The information loss thanks to quantization and random masking in FedVS outweighs that from localized training in DVFL. As a result, performance lags even in perfect conditions (Table \ref{tab:vfres_main}). We also see that DVFL performance tracks SplitNN's closely on all datasets, including tabular ones which are typically ``hard" for neural networks.
\begin{table*}[] 
\centering
\caption{On a variety of typical vertically partitioned datasets, we see that DVFL performs comparably to SplitNN even under perfect conditions. Even on tabular datasets harder (especially for NNs) than MNIST such as Credit Card and Parkinson's, DVFL model performance is comparable to SplitNN. Despite using a smaller model, both outperform the results cited in the FedVS paper \citep{FedVS}.}
\vspace{2ex}
\label{tab:vfres_main}
\resizebox{\linewidth}{!}{%
\begin{tabular}{@{}ccccccc@{}}
\toprule
\multirow{3}{*}{\textsc{\textbf{Strategy}}} & \multicolumn{4}{c}{\textsc{\textbf{Test Accuracy (\%)}}} &  \multicolumn{2}{c}{\textsc{\textbf{Desiredata}}}          \\ \cmidrule(){2-5} \cmidrule(){6-7} 
                          & \multicolumn{4}{c}{\textsc{Dataset}}& \multirow{2}{*}{\textsl{Privacy}}&    \multirow{2}{*}{\textsl{Fault Tolerance}}          \\ \cmidrule(lr){2-5} 
                          & \textsl{Credit Card} & \textsl{Parkinson's} & \textsl{CalTech-7} & \textsl{Handwritten} && \\ \midrule
DVFL                      & 82.66       &85.76             & 99.00          & 98.25    & No Gradients & Crash   \\
VFL (SplitNN)                   & 84.11    &86.64             & 99.66          & 98.25  & None & None          \\ 
FedVS \citep{FedVS}\tablefootnote{Results extracted from graphs. Raw figures were unavailable and the authors did not respond to our inquiries.} & 81.99 & 81.83 & 95.98 & 98.25 & Quantization and Noise & Straggler
\\\bottomrule
\end{tabular}%
}
\end{table*}
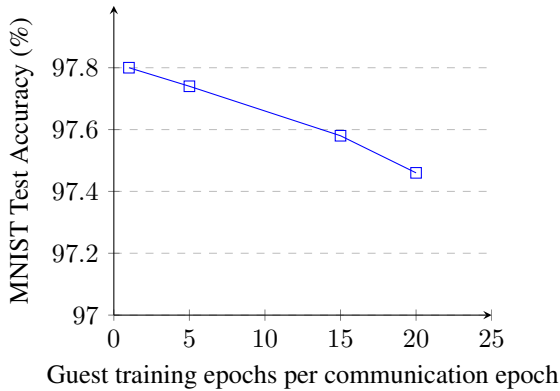
\begin{figure}[H]
\centering
\begin{tikzpicture}
\begin{axis}[
    axis lines = left,
    xlabel={Guest training epochs per communication epoch},
    ylabel={MNIST Test Accuracy (\%)},
    xmin=0, xmax=25,
    ymin=97.0, ymax=98.0,
    xtick={0,5,10,15,20,25},
    ytick={97.00,97.20,97.40,97.60,97.80},
    legend pos=south west,
    ymajorgrids=true,
    grid style=dashed,
    width=0.8\columnwidth 
]
\addplot[
    color=blue,
    mark=square,
    ]
    coordinates {
    (1,97.80)(5,97.74)(15,97.58)(20,97.46)
    };
\end{axis}
\end{tikzpicture}
\caption{Model performance degrades with an increase in communication period, i.e. more communication is correlated with better performance.}
\label{fig:comm_main}
\end{figure}
\paragraph{Comparison to FedOnce}
FedOnce \citep{Wu2022} aims to reduce the volume of communication at the cost of model performance. Equivalently, \citet{pmlr-v119-belilovsky20a, siddiqui2023blockwise} demonstrate that updating all layers on local objectives on each forward pass performs better than fully training and freezing a layer before moving on to the next. Even if we consider this completely fault tolerant, the overall performance loss from reduced communication is enough that DVFL performs better or comparably even in our experiments with the high fault rates. If communication size is a concern, DVFL provides fine-grained control over that. This is discussed further in \S\ref{sec:CommVol}.

\subsection{Redundancy Improves Performance, Asynchronicity Affords Efficiency}
We observe (Figure \ref{fig:faults}) that test accuracy improves with the number of hosts with all other system parameters fixed. This is the case even when there are no faults. This is not surprising, as the owner's aggregation of host results can be seen as a form of bagging \citep{Breiman1996}. As a corollary, the degradation of performance with connection faults is more graceful with increasing redundancy.

\subsection{Out-of-Intersection Training and Privacy}
Although VFL with SplitNN marginally outperforms our DVFL NN in the experiments with perfect conditions, DVFL significantly outperforms VFL on a limited labeled intersection. DVFL's advantage is greater with fewer labels (Table \ref{tab:labeled}). DVFL is performant while also completely eliminating the possibility of gradient-based inference attacks. As reviewed in \ref{sec:related}, complex mechanisms can be developed on top of SplitNN to provide privacy guarantees within some bounds. DVFL achieves complete privacy against gradient-based attacks without any extra computation and minimal performance loss.
\begin{table}[]
\caption{We exploit data outside the sample intersection to train DVFL's guests and hosts. For the owner model, we use labeled entities common to all datasets. This performs better than VFL, which can only use the common samples.}
\vspace{5ex}
\centering
\label{tab:labeled}
\resizebox{!}{!}{%
\begin{tabular}{@{}ccccc@{}}
\toprule
\multirow{3}{*}{\textsc{\textbf{Strategy}}} & \multicolumn{4}{c}{\textsc{\textbf{Test Accuracy (\%)}}}    \\ \cmidrule(l){2-5} 
                          & \multicolumn{4}{c}{\textsc{\# of Labeled Samples}} \\ \cmidrule(l){2-5} 
                          & \textsl{128}       & \textsl{256}      & \textsl{512}      & \textsl{1024}    \\ \toprule
DVFL                      & 76.45     & 82.51    & 85.18    & 88.90   \\
VFL (SplitNN)             & 65.07     & 77.51    & 83.09    & 87.26   \\ \bottomrule
\end{tabular}%
}
\end{table}
\section{Engineering Flexibility with DVFL}
\subsection{Control over Communication Volume}
\label{sec:CommVol}
Since guests and hosts both train on their own objectives, it is not necessary that they synchronize on every iteration. This means communication cost can be controlled by only transferring some embeddings, at the cost of performance. We experiment with various communication frequencies and present the results in Table \ref{fig:comm_main}. These experiments are discussed in further detail in the Appendix \ref{appendix:comm}.



\subsection{Control over Computational Cost}
\label{sec:CompCost}
Ostensibly, DVFL is slightly more computationally expensive than VFL since both guests and hosts must generate their own update signals in the absence of supervisory labels. In our experiments with autoencoders, this manifests as training the decoder head. However, DVFL allows the flexibility of allowing each participant to train for a different number of iterations. This is unlike VFL where the guest must compute a forward pass and then update every time the host makes a request. As such, the number of iterations for each participant can be tuned to balance model performance and computational cost. 

Our main experiments with DVFL only use 20 epochs of guest training while VFL hosts, DVFL hosts and DVFL owners use 60 epochs. We measure the wall-clock time of each process in the Appendix \ref{appendix:time} and see that the guest processes are actually alive for a shorter duration in DVFL in comparison to VFL ($\approx 55.88$s vs $\approx 80.22$s under the same operating conditions). Indeed, the same way that the meaningful learning is possible even if the guest sends a limited number of activations, the host may make do with storing a limited number of activations. In the Appendix \ref{appendix:space} we discuss how a memory constrained host may still meaningfully learn while only storing the last few activations rather than the entire history.
\section{Limitations}
\label{sec:lim}
Our models are shallow. \citet{wang2021revisiting} show that the performance improves with the length of backpropagation in blockwise-trained networks, and suggest that it is due to increasing mutual information between the shallower and deeper layers. Further, given a model, the 
loss does not guide the model to extract strictly task-useful features. These issues typically also lead to scalability issues to harder datasets, although we observe in our experiments on tabular datasets that the difference is not that large. However, we expect that the information loss from the absence of direct connections between guests' neurons is greater than that due to the loss of task-relevant information.

\section{Broader Impact} We believe DVFL will enable large scale and democratic participation in joint learning tasks where VFL has been unable to provide solutions. e.g. street \& private residential cameras predicting traffic incidents at a busy intersection. Participants might not all be able to see each training incident i.e. small $\mathcal S_\text{guest}\cap S_\text{labels}$, or might be concerned about privacy, or suffer from intermittent connection. DVFL is also more practical for existing cross-silo applications. Consider the typical example of various specialized healthcare providers predicting, say, cancer. Not only can the hospitals train asynchronously and not worry about faults, they need not worry about compromised hospitals launching gradient inference attacks. Further, they may now admit more hospitals to their system even if they have few patients in common. We see DVFL as a step towards usable and ubiquitous machine learning.
\section{Conclusion and Future Work}
We present Decoupled Vertical Federated Learning, a strategy for joint learning on vertically partitioned data. Instead of adding complexity to cope with the data leakage and synchronization associated with BP, we eschew feedback altogether. Combined with the ability to reuse feedforward signals, this allows for fault tolerance in training.

All-in-all, we address three impracticalities with VFL training: single points of failure, training data being limited to the intersection, and gradient-based privacy attacks: Training may continue even when there are faults, or there are a limited number of samples common to all guests. Malicious or compromised guests can no longer use gradient feedback to figure out possibly sensitive information. Existing techniques to enforce privacy guarantees either hamper model convergence, are expensive to implement or both.

In turn, DVFL introduces new flexibility and practicality to systems intended for learning from vertically partitioned data-- the volume of communication, memory usage and overall complexity can be controlled by a designer.

 Our results suggest DVFL's performance to be comparable to SplitNN on various ML tasks. Unlike SplitNN, DVFL degrades gracefully with manykinds of loss faults. Even with explicit fault handling strategies, SplitNN's performance drops off more sharply than DVFL's. In general, DVFL's asynchronicity provides a great deal of engineering flexibility. Some of it is discussed in the Appendix.

The next step for DVFL is to generalize it for more tasks and general architectures. VFL is readily compatible with a variety of machine learning models (including ensemble models). With a judicious choice of training objective for guest models, DVFL may also be extended past NNs.

\nocite{langley00}

\bibliography{example_paper}
\bibliographystyle{mlsys2024}


\newpage
\appendix
\clearpage

\section{DVFL for Split Neural Network Training}
\label{appendix:dvflnn}
\vspace{-1.5ex}
\begin{algorithm}[H]
\small 
\caption{DVFL training for split neural networks (DVFL-NN)}
\textsl{Input:} Guest datasets $\mathcal D_i$, optimizers $\text{optim}_{g_i}$ $\forall i\in$ $\{o\} \cup$ $\{g_1,\dots,g_{|\mathcal{G}|}\}\cup\{h_1,\dots, h_{|\mathcal H|}\}$\text{, communication schedule}\\
\textbf{Guest Training Round:}\\
\begin{algorithmic}[1]
\label{alg:DVFL}
  \FOR{each guest $g_i$ in parallel}
    \STATE Take the next batch $\mathbf x_{t, \mathcal F_i}$ from dataset $\mathcal D_i$
    \STATE $\hat{\mathbf{x}}_{t, i}\gets m_i(\mathbf x_{t,\mathcal F_i};\theta_i)$ \COMMENT{Guest forward pass}
    \IF{communication round}
    \FOR{each host $h_j\in \mathcal H$}
    \STATE $\mathbf B_{i, j}\gets \hat{\mathbf{x}}_{t, i}$
    \ENDFOR
    \ENDIF
    \STATE Compute an unsupervised loss $\mathcal L_i(\hat{\mathbf{x}}_{t, i})$
    \STATE $\theta_i\gets \text{optim}_i(\theta_i, \nabla_{\theta_i} \mathcal L_i)$ \COMMENT{Update guest model}
  \ENDFOR
\end{algorithmic}
\textbf{Host Training Round:}
\begin{algorithmic}[1]
  \FOR{each host $h_i$ in parallel}
    \IF{communication round}
    \STATE Read $\mathbf{B}_{j, i} \forall j\in\{1, 2,\dots,|\mathcal{G}|\}$
    \STATE Concatenate all $\mathbf B_{j,i}$ and append to $\mathbf A_i$
    \ENDIF
    \STATE Read vector $\mathbf b_{t, i}$ from $\mathbf{A}_i$
    \STATE $\hat{\mathbf{b}}_{t, i}\gets m_{h, i}(\mathbf b_{t,i};\theta_{h,i})$ \COMMENT{Host forward pass}
    \STATE Compute an unsupervised loss $\mathcal L_{h,i}\left(\hat{\mathbf{b}}_{t, i}\right)$
    \STATE $\theta_{h,i}\gets \text{optim}_{h_i}(\theta_{h_i}, \nabla_{\theta_{h_i}} \mathcal L_{h_i})$ \COMMENT{Update host model}
  \ENDFOR
\end{algorithmic}
\textbf{Owner executes:}
\begin{algorithmic}[1]
\REPEAT 
\STATE Guest Training Round
\STATE Host Training Round
\UNTIL convergence or a fixed number of iterations
\REPEAT
  \FOR{each entity $\mathbf{x_j}$, $j \in \mathcal{S}_{\mathrm{guests}} \cap \mathcal{S}_\mathrm{labels}$}
    \FOR{all guests in parallel}
      \STATE Get features $\mathbf x_{j, \mathcal F_i}$ pertaining to entity $\mathbf{x_j}$
      \STATE Pass features through guest models and send activations to hosts
    \ENDFOR
    \FOR{all hosts in parallel}
      \STATE Concatenate all received activations
      \STATE Pass activations through host models $m_{h, i}$ with parameters $\mathbf{\theta}_{h, i}$
      \STATE Send all activations to the owner
    \ENDFOR
    \STATE Owner concatenates all inputs to form $\mathbf x_{j,\mathrm{enc}}$
    \STATE $\hat{\mathbf{y}}_{j}\gets m_o({\mathbf x}_{j,\mathrm{enc}};\theta_o)$ \COMMENT{Owner forward pass}
    \STATE Compute loss against labels $\mathcal L_o(\hat{\mathbf y}_j, \mathbf y_j)$
    \STATE $\theta_i\gets \text{optim}_o(\theta_o, \nabla_{\theta_o} \mathcal L_o)$ \COMMENT{Update owner model}
  \ENDFOR
\UNTIL convergence or a fixed number of iterations
\end{algorithmic}
\end{algorithm}

\vspace{-5ex}
\section{Results on More Datasets, Comparison to FedVS}
\label{appendix:moredata}
In order to bolster the comparison to SplitNN, we evaluate the performance of DVFL on some benchmark vertically partitioned datasets. They are Credit Card \citep{YEH20092473}, Parkinson's \citep{Sakar2019ACA}, CalTech-7 and Handwritten \citep{uci}. Note that CalTech-7 is a subset of 7 classes CalTech 101 \citep{li_andreeto_ranzato_perona_2022}: Face, Motorbikes, Dolla-Bill, Garfield, Snoopy, Stop-Sign and Windsor-Chair. Both Handwritten and CalTech have 6 distinct ``views"-- features of distinct domains, e.g. spatial, frequency (Fourier), Gabor filter coefficients, etc.

FedVS \citep{FedVS} is a state-of-the-art VFL algorithm that aims to tackle both privacy and fault tolerance. There is a round of secret shares so that each guest holds an encrypted version of a horizontal section of other guests' data and model parameters. The shares are arranged such that if $\leq L$ (a fixed system parameter) guests don't send their outputs at the forward pass, there is enough redundant information in the other guests' forward passes to account for it. In order to maintain privacy despite this sharing of data, guests quantize their models and data and apply randomly mask. This means there is no longer enough information for a malicious guest to exactly reconstruct the data it receives from other guests. But, this obscurity is at the cost of model performance. Further, this algorithm is contingent on the guests using polynomial networks \citep{livni2014algorithm, polynet}.

In the following experiment, we compare our DVFL algorithm, SplitNN and FedVS. Our VFL and DVFL models are of the same depth as FedVS, but we use a smaller width. We also only use population normalization (as opposed to online batch and layer normalization). Therefore, the non-FedVS models have much fewer parameters and are simpler to train. We also take the liberty of using non-polynomial activations (Leaky ReLU). Model details and hyperparameters are in the Appendix. $|\mathcal{H}| = 1$ for these experiments.

\begin{table}[] 
\centering
\caption{On a variety of typical vertically partitioned datasets, we see that DVFL performs comparably to SplitNN even under perfect conditions. Even on tabular datasets harder (especially for NNs) than MNIST such as Credit Card and Parkinson's, DVFL model performance is comparable to SplitNN. Despite using a smaller model, both outperform the results cited in the FedVS paper \citep{FedVS}. DVFL achieves this while tolerating a broader class of faults and also providing complete immunity to gradient-based attacks.}
\vspace{2ex}
\label{tab:vfres}
\resizebox{\columnwidth}{!}{%
\begin{tabular}{@{}ccccc@{}}
\toprule
\multirow{3}{*}{\textsc{\textbf{Strategy}}} & \multicolumn{4}{c}{\textsc{\textbf{Test Accuracy (\%)}}}              \\ \cmidrule(){2-5} 
                          & \multicolumn{4}{c}{\textsc{Dataset}}                         \\ \cmidrule(lr){2-5} 
                          & \textsl{Credit Card} & \textsl{Parkinson's} & \textsl{CalTech-7} & \textsl{Handwritten} \\ \midrule
DVFL                      & 82.66       &85.76             & 99.00          & 98.25       \\
VFL (SplitNN)                   & 84.11    &86.64             & 99.66          & 98.25            \\ 
FedVS \citep{FedVS}\tablefootnote{Results extracted from graphs. Raw figures were unavailable and the authors did not respond to our inquiries.} & 81.99 & 81.83 & 95.98 & 98.25
\\\bottomrule
\end{tabular}%
}
\vspace{-2ex}
\end{table}

The feature splits and train-test splits for each dataset are identical to the FedVS paper \citep{FedVS}. 
Note that the datasets here are smaller than MNIST (30000 samples for credit card, 756 for Parkinsons, 1474 for MNIST, 2000 for handwritten). Observe that VFL and DVFL both outperform FedVS, which is quantized and constrained only to use polynomial networks \citep{livni2014algorithm}.

It is not meaningful to compare these methods under faults, since they are designed for different fault models. FedVS is completely tolerant to stragglers, particularly the case where stragglers are bounded in number. Like FedVS, this fault model does not affect DVFL, since guests and hosts train independently to each other. Conversely, our fault model might cause a catastrophic failure to FedVS: an arbitrary number of participants may crash, which is as if there is a number of stragglers greater than the predetermined system parameter.
\section{Other Failure Modes}
\label{ofm}
Here, we have considered only crash faults. Other typical failure modes for distributed systems include arbitrary delays, incorrect order, and Byzantine faults. 
\vspace{-3ex}\paragraph{Arbitrary Delays and Incorrect Order} As discussed in the previous section and \ref{sec:related}, a line of work exists to partially tackle this problem. Typically, methods aimed at handling stragglers either demand a bound on the delay of stragglers or the number thereof. In return, methods such as FedVS \citep{FedVS} can train with no performance loss. On the other hand, with bookkeeping like our VFL variants from the main paper, we may observe graceful degradation with stragglers and incorrect order (or no degradation at all in the case of \emph{wait}, although that also requires a bounded delay). 
\vspace{-3ex}\paragraph{Byzantine Faults} Byzantine faults are the most severe fault model, where messages may be corrupted. A possible future direction for DVFL is to incorporate tolerance to Byzantine faults. Observe that our SplitNN-buffer strategy is, in a sense, a Byzantine fault, and it is not performant. Byzantine faults are not particularly difficult for DVFL during the guest-host phase, since the learning target is not task-relevant. But some investigation is required for Byzantine faults during the owner learning phase, which cannot be outright detected as crash faults can. 
\section{Models, Hyperparameters and Implementation Details}
\label{appendix:model}
\subsection{Models}
The ``global", single-participant neural network (see Figure \ref{fig:dvfldiagram}) being split is a shallow multi-layer perceptron. It is parameterized by $W_g$ and $W_h$. $W_g$ is the dimension of the concatenated output of all guests viz. the input to a host. $W_h$ is the output dimension of a single host. 

\subsubsection{MNIST}
Our batch sizes are 64.

\textbf{The ``guest" models} have 2 layers, which squeeze the flattened $\frac{784}{|\mathcal{G}|}$-long input vector dimension to $\frac{400}{|\mathcal{G}|}$ via LeakyReLU, and then to $\frac{W_g}{|\mathcal{G}|}$ with ReLU activation. 

\textbf{The ``host" models} squeeze the $W_g$-dimensional input to $\frac{W_g + 3W_h}4$ and then to $W_h$ via LeakyReLU, with Leaky ReLU activation (with negative half-plane slope = 0.01). 

\textbf{The ``owner" model} squeezes the $W_h\times |\mathcal{H}|$ dimensional input to $W_h$ and then to $40$ with Leaky ReLU activations and finally to a $10$-class prediction with an implicit softmax via the cross entropy loss function.

For DVFL experiments, the guest, host, and owner modules are separate objects simulated to reside on separate devices. In order to produce a reconstruction, they also contain decoder MLPs which undo the squeezing operation. The guest decoder is activated by a sigmoid function (since the input image is normalized to the range $[0,1]$) and the host decoder is activated by a ReLU function (since the output of the guest model is also activated by a ReLU function and therefore cannot be negative). 

For SplitNN experiments, the guest encoder architecture is identical to DVFL. The host model is a sequential module of the DVFL host encoder followed by the owner classifier.

We try DVFL for $W_g \in \{400, 320, 240, 200\}$ and $W_h\in \{200, 160, 120\}$. The classification accuracies are presented in Table \ref{tab:modelwidth}.

\begin{table}[H]
\centering
\caption{Ablation study of model width (particularly, the dimensions of the outputs which are communicated) on model performance with DVFL. Our experiments in the main paper use $W_g = 320$ and $W_h = 160$. We run a hyperparameter sweep for each cell.}
\vspace{2ex}
\label{tab:modelwidth}
\resizebox{\columnwidth}{!}{%
\begin{tabular}{@{}ccccc@{}}
\toprule
\multirow{2}{*}{$W_h$} & \multicolumn{4}{c}{\textsc{\textbf{Test Accuracy (\%)}}}                \\ \cmidrule(l){2-5} 
                       & $W_g = 200$ & $W_g = 240$ & $W_g = 320$ & $W_g = 400$ \\ \midrule
120                    & 97.43       & 97.28       & 97.09       & 97.34       \\
160                    & 97.60       & 97.86       & 97.89       & 97.69       \\
200                    & 97.81       & 97.89       & 98.15       & 98.30       \\ \bottomrule
\end{tabular}%
}
\end{table}

There two competing factors: the number of parameters (a monotonic function of $W_g\times W_h$) and the compression from host to guest. While more parameters implies a more generalizable (and hence hopefully accurate) model, an increase in $W_g$ for a fixed $W_h$ seems to suffer from some information loss, especially for larger $W_h$. It could be that the model eschews an encoding with more classification-relevant information when encodings with more reconstruction-relevant information are available thanks to the larger parameter space.

In the main paper, we use $W_g = 320$ and $W_h = 160$ for our experiments on MNIST. This choice enables good model performance while also demonstrating host models' ability to compress the data. 
\subsubsection{VFL Datasets}
Similar to MNIST, we use shallow multi-layer perceptrons for our results on vertically partitioned data in the Appendix. The model sizes for the guest $g_i$, host $h_k$ and owner $c$ are summarized in table \ref{tab:moremodels}, where $d_i$ is the input dimension of guest $g_i$.

\begin{table}[H]
\centering
\caption{Model dimensions used in our experiments on tabular datasets. The dimension of the input to a guest $g_i$'s model is $d_i$. $W_g$ is the dimension of the concatenated output of the guests' models. $W_h$ is the dimension of the output of one host's model.}
\vspace{1ex}
\label{tab:moremodels}
\resizebox{\columnwidth}{!}{%
\begin{tabular}{@{}ccccc@{}}
\toprule
\multirow{2}{*}{\textsc{\textbf{Dataset}}} & \multirow{2}{*}{$W_g$} & \multirow{2}{*}{$W_h$} & \multicolumn{2}{c}{\textsc{\textbf{Activation}}} \\ \cmidrule(l){4-5} 
            &          &                 & \textsc{Guest}      & \textsc{Host}       \\ \midrule
Credit Card & $\sum_i\left\lceil\frac{3d_i}{4}\right\rceil$       & 10             & Leaky ReLU & Leaky ReLU \\
Parkinsons  & $\sum_i\left\lfloor\frac{d_i}{4}\right\rfloor$    & $\frac{\sum_i\left\lfloor\frac{d_i}{4}\right\rfloor}{4}$          & Leaky ReLU & ReLU       \\
CalTech-7   & $256\times|\mathcal{G}|$       & 256              & Leaky ReLU & ReLU       \\
Handwritten & $\sum_i\left\lceil\frac{3d_i}{4}\right\rceil$ &$3\frac{\sum_i\left\lceil\frac{3d_i}{4}\right\rceil}{4}$ & Leaky ReLU & ReLU       \\ \bottomrule
\end{tabular}%
}
\end{table}

\subsection{Fault Simulation}
\label{appendix:faults}
We simulate faults based on six hyperparameters:
\begin{enumerate}
    \vspace{-2ex}
    \item \textbf{Connection Faults}: During each connection round, for each pair $(g_j, h_i)$ we draw a sample from a uniform distribution with support $[0,1]$. We track the current status of each connection using flags.
    \begin{itemize}
        \item $R^{(d)}_\mathrm{connection}$: If the connection between $(g_j, h_i)$ was alive on the last iteration, we kill it if the sampled value is greater than $R^{(u)}_\mathrm{connection}$. 
        \item $R^{(u)}_\mathrm{connection}$: If the connection between $(g_j, h_i)$ was dead on the last iteration, we revive it if the sampled value is lesser than $R^{(u)}_\mathrm{connection}$.
    \end{itemize}
    In DVFL, we only write $g_j$'s activations to register $\mathbf B_{j,i}$ if. Similarly, for SplitNN, we check and update the flag before any forward pass from or backward pass to the split layer.
    
    \vspace{-1ex}
    \item  \textbf{Guest Faults}: Whenever a guest is called (i.e. there is a forward or backward pass), for each $g_j$ we draw a sample from a uniform distribution with support $[0,1]$.
    \begin{itemize}
        \item $R^{(d)}_\mathrm{guest}$: If the guest $g_j$ was alive on the last call, we kill it if the sampled value is greater than $R^{(u)}_\mathrm{guest}$. 
        \item $R^{(u)}_\mathrm{guest}$: If the guest $g_j$ was dead on the last call, we revive it if the sampled value is lesser than $R^{(u)}_\mathrm{guest}$.
    \end{itemize}
    In DVFL, we only update $\theta_j$ if the guest $g_j$ is alive. If it is a connection epoch, we only write $g_j$'s activations to $\mathbf B_{j,i} \forall i\in\{1,2,\dots,|\mathcal{H}|\}$ if the guest is alive. Similarly, in SplitNN, we only allow a guest to send a forward pass and receive gradients if it is alive.
    \vspace{-1ex}
    \item \textbf{Host Faults}: During each training iteration, for each $h_i$ we draw a sample from a uniform distribution with support $[0,1]$.
    \begin{itemize}
        \item $R^{(d)}_\mathrm{host}$: If the guest $h_j$ was alive on the last call, we kill it if the sampled value is greater than $R^{(u)}_\mathrm{host}$. 
        \item $R^{(u)}_\mathrm{host}$: If the guest $h_j$ was dead on the last call, we revive it if the sampled value is lesser than $R^{(u)}_\mathrm{host}$.
    \end{itemize}
    In DVFL, we only update $\theta_{H,i}$ if the host is alive. If it is a communication epoch, we also drop the last item in input buffer $\mathbf A_{i}$. In the case of SplitNN, we assume the host is able to call on the guests before failing. When a SplitNN host fails, it does not compute any gradients.
    \vspace{-1ex}
\end{enumerate}
\subsection{Other Hyperparameters}
\label{appendix:hyperparams}
For all experiments, we perform a hyperparameter search. The values we used for the results presented in this paper are below.
\subsubsection{MNIST}
For guests and DVFL hosts, we use an Adam optimizer with the default $\beta_1 = 0.9, \beta_2 = 0.999$. We use a learning rate of $2\times10^{-3}$ (guest, host) and a weight decay of $10^{-5}$. For DVFL owners, we use SGD with a learning rate of $8\times10^{-2}$. For SplitNN guests, we use stochastic gradient descent with a learning rate of $10^{-2}$ and momentum of $0.5$. In general, guests were trained for 20 epochs, hosts for 40 and owners for 60. For SplitNN, the whole model was trained end-to-end with 60 epochs. For the experiments on limited intersecting datasets, owners/SplitNN hosts are trained for 160 epochs. In all cases, DVFL owner models and SplitNN host models were trained with early stopping.
\subsubsection{VFL Datasets}
Hyperparameters are presented in Table \ref{morehyper}.

\begin{table*}[]
\centering
\caption{Hyperparameters for the results presented in the Appendix.}
\resizebox{\textwidth}{!}{%
\begin{tabular}{@{}cccccc@{}}
\toprule
\multirow{4}{*}{\textsc{\textbf{Dataset}}} & \multicolumn{5}{c}{\textsc{\textbf{Learning Rate}}}                                                                                                                                     \\ \cmidrule(l){2-6} 
                         & \multicolumn{3}{c}{\textsc{DVFL}}                                                                      & \multicolumn{2}{c}{\textsc{SplitNN}}                                           \\ \cmidrule(lr){2-4}\cmidrule(lr){5-6} 
                         & \textsl{Guest}                             & \textsl{Host}                              & \textsl{Owner}                 & \textsl{Guest}                             & \textsl{Host}                              \\
                         & (Adam, $\beta_1$ = 0.9, $\beta_2$ = 0.999) & (Adam, $\beta_1$ = 0.9, $\beta_2$ = 0.999) & (SGD, momentum = 0.5) & (Adam, $\beta_1$ = 0.9, $\beta_2$ = 0.999) & (Adam, $\beta_1$ = 0.9, $\beta_2$ = 0.999) \\ \midrule
Handwritten              & 2e-3                              & 1e-3                              & 3e-3                  & 1e-3                              & 1e-4                              \\
CalTech-7                & 5e-4                              & 1e-2                              & 3e-5                  & 1e-4                              & 1e-4                              \\
Parkinson's              & 5e-3                              & 4e-3                              & 5e-4                  & 1e-4                              & 5e-5                              \\
Credit Card              & 1e-3                              & 1e-3                              & 1e-3                  & 1e-3                              & 1e-3                              \\ \bottomrule
\label{morehyper}
\end{tabular}%
}
\end{table*}
\subsection{Implementation Details}
We simulate parallel training by time multiplexing and using preallocated replay and communication tensors. The experiments were run on a variety of hardware, including mobile CPUs (Apple Silicon M2) and workstation GPUs (on a single NVIDIA RTX A5000 and on a single NVIDIA Tesla K40). 
\section{More Discussion on Related Works}
\label{appendix:rw}
\subsection{Fault-Tolerant Distributed Optimization}
The impact of redundancy for fault tolerance in distributed systems is well studied. \citet{Liu2021} provide a rigorous analysis in the case of distributed optimization, showing  that systems which are $2f$-redundant are approximately resilient to $f$ Byzantine agents\if (within bounds)\fi. The authors also present two fault-tolerant algorithms \if to handle faults\fi for horizontally distributed convex optimization. \citet{liu2021survey,bouhata2022byzantine} extensively review approaches to fault tolerance in distributed learning. Most methods \if depend on\fi require synchronization by an aggregator and incur computational overhead. Further, most methods involve manipulating gradients, which forces a choice between convergence rate and tightness. 

\textbf{For VFL}: A limited number of works exist to tackle the related problem of straggling participants. 

\citet{FedVS} introduce FedVS, an algorithm for systems where guest models are polynomial networks \citep{livni2014algorithm,polynet}. FedVS uses a form of secret sharing to introduce redundancy in the input data. The redundancy is so arranged such that, when there are at most a fixed number of failed or late guests, timely guests have enough combined information to account for the missing messages. 

In situations where multiple participants might have labels (i.e. there is a horizontal federation of label owners), \citep{CVFL} exploits redundancy and periodic synchronization to provide straggler resistance. One specially designated horizontally federated host is responsible for orchestrating the synchronization between hosts. Typically, straggler-resistant algorithms are either designed to tolerate a fixed number of stragglers or a finite duration of lag (i.e. a crash would still be catastrophic). Further, FedVS and its predecessors \citep{Shi2022, chen2020vafl, Hu2019} typically only address faults in the forward pass.


\subsection{Federated Transfer Learning}
Federated Transfer Learning (FTL) is a strategy for systems where there is some overlap in both sample space and feature space. The host owns labels and some features (i.e. it is an active participant) for entities in its sample space. Guests have limited labels for their task, but have some small overlap in feature space and sample space with the host (``aligned" samples).

\citet{Liu2020} present Secure Transfer Federated Learning (STFL), a framework for a single host and a single guest. Both parties learn a common latent space using local feature extractors. Encodings of the passive participant’s unaligned samples in this feature space is used by the passive participant to learn its task model. \citet{feng2020multiparticipant,10.1016/j.knosys.2022.109384} extend this idea to multiple parties.

Much like DVFL, these methods exploit an unsupervised latent space to learn from data that is not present in the intersection. However, unlike DVFL, these methods exploit the existence of common feature space between participants. This problem is relevant to robust distributed sensor fusion \citep{liu2021survey}, where there is some redundancy/overlap in the ``view" of each participant. 

For the related case of VFL with small $|\mathcal{S}_{\mathrm{guests}}\cap \mathcal{S}_\mathrm{labels}|$, \citet{Sun_2023_ICCV} introduce few-shot Vertical Federated Learning, where \if one (or a small number of)\fi a few epochs of training on the aligned dataset are followed by semi-supervised learning on the unlabelled entities of each client's dataset. \if In order to\fi To generate labels\if for semi-supervised learning\fi, guests cluster the gradients they receive from the aligned samples. While \if the authors demonstrate \fi this enables remarkable model performance, there is unwanted computational overhead required to generate the labels. Moreover, this \if approach\fi is a form of label inference attack! Not only is this a breach of privacy, this clustering could be used downstream by a malicious agent to perform feature inference attacks (see \S\ref{sec:Inference}).
\subsection{Greedy and Localized Learning}

Algorithms which update parameters in hidden layers of NNs based on local or auxiliary objectives (as opposed to the global label loss) have recently seen renewed interest. In the scope of VFL training, they enable some useful properties for NN training.

The primary drawbacks of BP in the scope of VFL are the inherent locking synchronization constraints, described in \citet{pmlr-v70-jaderberg17a}. In order to update its weights, a layer must wait for the backward signal to reach from the output layer (backward locking). In order to begin updating any layer, a full forward pass must first propagate to the end of the network (update locking). In order for a layer to evaluate an input to begin with, a forward pass must propagate to it (forward locking). These strict locks at the split layer are exactly the failure points for VFL training-- forward locking causes training to fail if a guest does not communicate activations, backward locking causes training to fail if the host does not communicate gradients. 

By eschewing the wait to evaluate the global objective, most greedy algorithms are backward unlocked if not update unlocked.
\citet{pmlr-v119-belilovsky20a} propose a replay buffer mechanism to partially address forward locking for model parallel training. Each agent responsible for training a module uses a replay buffer to store activations from the previous module's agent. During a forward pass, the agent reads an activation from the buffer. If the forward signal has propagated through the previous module, it would have written the latest activation to the buffer. If not, the previous activation is reused.

\section{Communication and Space Complexity}
\label{appendix:complexity}
\subsection{Periodic Communication}
\label{appendix:comm}
The owner initiates each round of training as well as communication for all agents. Since the training of guests and hosts are decoupled, the owner may ask guests to communicate their latent activations (and therefore start host training rounds) at arbitrary intervals. In VFL, guest model updates are dependent on host feedback. So there is every-shot communication: guests send every single output to the host. 

With one-shot communication such as FedOnce \citep{Wu2022}, the volume of communication is greatly reduced, at the cost of model performance. Equivalently, \citet{pmlr-v119-belilovsky20a, siddiqui2023blockwise} demonstrate that updating all layers on local objectives on each forward pass performs better than fully training and freezing a layer before moving on to the next. 

We study the tradeoff between communication cost and model performance by introducing a hyperparameter $K$, which is the period (in guest epochs) at which \emph{communication epochs} occur. During a communication epoch, every iteration of the guest model iteration is followed by a communication round i.e. guests write their activations to the communication registers, and hosts read them, concatenate them and store the concatenations in their input buffers.\\If there is a host training iteration at every guest training iteration communication round, after the guests complete $N$ epochs, the hosts will have
\begin{enumerate}
\vspace{-2ex}
    \item A model trained for $\left\lfloor \frac NK \right\rfloor$ epochs
    \vspace{-2ex}
    \item The history of model inputs from each iteration in $\left\lfloor \frac NK \right\rfloor$ epochs of training
    \vspace{-2ex}
\end{enumerate}
If the host must complete a certain number $M > \frac{|\mathcal{S}_{\mathrm{guests}}|}{\text{batch size}}\times \left\lfloor \frac NK \right\rfloor$ iterations of training, it reuses these activations until the required number of iterations is met.

We simulate VFL training with $4$ hosts and $0$ fault rate. We measure test accuracy for the following communication periods:
\begin{itemize}
\vspace{-3ex}
    \item Every guest epoch ($K = 1$)
    \vspace*{-2ex}
    \item Every 5 guest epochs ($K = 5$)
    \vspace*{-2ex}
    \item Every 10 guest epochs ($K = 10$)
    \vspace*{-2ex}
    \item Only once, after 20 guest epochs ($K = 20$).
    \vspace*{-2ex}
\end{itemize}
Note that the dimension of the latent representations communicated by each guest to reach host is $80$, and each dimension corresponds to a 32-bit floating point number. The MNIST training dataset has $60000$ samples. If the total number of guest training epochs is $N$ and the period of communication epochs is $K$, the total communication cost per guest is $|\mathcal{H}|\times\lfloor \frac NK\rfloor \times 60000\times80\times32$ bits. Each message received is saved by the host in the activation replay\footnote{Although this is not necessary if the host will not reuse activations, viz. a host wants to train for $k$ iterations and it receives activations of $k$ iterations from the guests. This is discussed in more detail in the Appendix \ref{appendix:space}}.
\begin{figure}[H]
\centering
\begin{tikzpicture}
\begin{axis}[
    axis lines = left,
    xlabel={Guest training epochs per communication epoch},
    ylabel={MNIST Test Accuracy (\%)},
    xmin=0, xmax=25,
    ymin=97.0, ymax=98.0,
    xtick={0,5,10,15,20,25},
    ytick={97.00,97.20,97.40,97.60,97.80},
    legend pos=south west,
    ymajorgrids=true,
    grid style=dashed,
    width=0.8\columnwidth 
]
\addplot[
    color=blue,
    mark=square,
    ]
    coordinates {
    (1,97.80)(5,97.74)(15,97.58)(20,97.46)
    };
\end{axis}
\end{tikzpicture}
\caption{Model performance degrades with an increase in communication period, i.e. more communication is correlated with better performance.}
\label{fig:comm}
\end{figure}
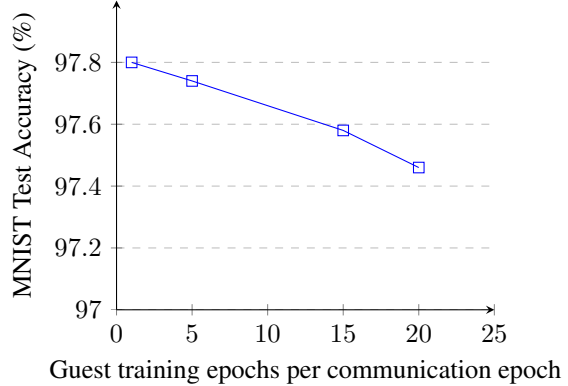

Consistent with results from \citet{pmlr-v119-belilovsky20a,Wu2022,siddiqui2023blockwise}, training the depth-wise split model parallely yields better model performance. We observe the tradeoff between communication cost and model performance. With 1.536GB, 0.307GB, 0.154GB and 0.077GB outgoing from each guest, the accuracies on the MNIST test set are $97.30\%$, $97.14\%$ $97.08\%$ and $96.84\%$ respectively.
\subsection{Activation Replay Mechanism \& its Computational Overhead}
\label{appendix:space}

A host gets its input from its activation replay buffer. That allows it to train asynchronously wrt the guests. The host can train independently of the guests as long as the guests have communicated earlier. Indeed, that is the case in our experiments-- we simulate host models' training offline, after guest training. As discussed in Appendix \ref{appendix:comm}, this allows control over when the guests communicate. 

Another way to look at this is: hosts do not need to train for the same number of iterations as guests communicate messages for. This includes the case where guests communicate every iteration, but hosts want to train more (or less) iterations. As we detailed in Appendix \ref{appendix:hyperparams}, we actually implement this for our experiments. While we use 40 epochs to train the entire SplitNN model, we only use 20 epochs to train the DVFL guests. Guests need not remain online for any more time than required to train 20 epochs. So, not only does this allow for more flexibility, it also helps fault tolerance.




\paragraph{Computational Resources} In order to maintain this asynchronicity, hosts need to maintain the activation replay buffer. For each message it receives, the size of the buffer grows by $\text{batch size}\times W_g$. Further, these expensive memory read/write operations also add time complexity to DVFL hosts' training. This overhead does not exist in standard VFL. This it is DVFL's primary drawback\footnote{The other source of overhead is the calculation of the self-supervised/unsupervised loss. For example, consider our experiments with MSE as the guest and host objectives. Along with our model, we need to train a decoder head that projects the guest output onto a space of the same dimension as input. Modern contrastive self-supervised losses e.g. \citet{SimCLR, barlow} typically use a shallow projector network from the desired representation onto a latent space. The loss is calculated using the projected vectors. Intuitively, in the absence of a pre-computed ground truth label, the model must perform extra computation to generate the ``self-"supervisory signals.}, although it can be mitigated to some extent since some models can be trained for fewer epochs as discussed above. However, it is important to note that it is not strictly necessary to save all activations. If a host trains for $\leq$ the number of epochs that guests train/communicate for, it does not need to reuse activations. In that case, a host model input can be deleted after it is used once.

In general, it is not necessary for any pair of participants (either belonging to any set: be it $\mathcal G$, $\mathcal H$ or $\{\text{owner}\}$) to train their model for the same number of iterations. This flexibility comes at the cost of space, but the upside from a system design and fault tolerance perspective is salient.

\subsection{Empirical Time Complexity}
\label{appendix:time}
We compare the following regimes:
\begin{enumerate}
    \vspace{-2ex}
    \item DVFL
    \begin{enumerate}
        \item MSE loss
        \begin{itemize}
            \item Same setup as main experiments
            \item Guests trained for 20 epochs
            \item Hosts trained for 40 epochs
            \item Owner trained for 60 epochs
        \end{itemize}
        \item SimCLR loss
        \begin{itemize}
            \item Guests trained for 5 epochs
            \item Owner trained for 60 epochs
            \item Guests optimized using Adam with $4\times10^{-3}$ learning rate, $\beta_1 = 0.9$, $\beta_2 = 0.999$, weight decay $10^{-5}$
            \item $\text{linear}(160,50)\to\text{batchnorm}\to\text{ReLU}\to\text{linear}(50,50)\to\text{batchnorm}\to\text{ReLU}\to\text{linear}(50,50)$ projector
            \item No host
            \item Owner model is equivalent to the host model and owner model of our main experiments in sequence
            \item Owner is optimized using SGD with learning rate $10^{-1}$, momentum 0.9
            \item Batch size of 600.
        \end{itemize}
    \end{enumerate}
    \vspace{5ex}
    \item VFL
    \begin{enumerate}
        \item No localized loss
        \begin{itemize}
            \item Same setup as main experiments
            \item System trained for 60 epochs
        \end{itemize}
    \end{enumerate}
\end{enumerate}
We measure the CPU process time on an Apple Silicon M2 CPU and compare the following process + system times:
\begin{itemize}
    \vspace{-1ex}\item Forward: time per epoch spent in the forward pass (including loss calculation, if any)
    \vspace{-1ex}\item Backward: time per epoch spent in the backward pass (including optimizer step)
    \vspace{-1ex}\item Wait: time per epoch spent waiting for gradients
    \vspace{-1ex}\item Process: total time the process must be alive
\end{itemize}

\vspace*{-2ex}

We do not include communication delay or spent managing tensors, as those are heavily implementation-dependent. We also do not count data augmentation time for SimCLR, since that may be precomputed before guest-host training. 

\begin{table}[]
\centering
\caption{DVFL allows guests to train without wait and terminate their processes much sooner than VFL. Note that SimCLR requires a large batch size (here: 600), and therefore gets swapped out of memory frequently during training. The time reported includes that.}
\label{tab:my-table}
\resizebox{\columnwidth}{!}{%
\begin{tabular}{lllll}
\cline{1-5}
                                               &                                                     & \multicolumn{3}{c}{\textsc{Strategy}}         \\ \cline{3-5}
\multicolumn{1}{c}{}                           &                                                     & DVFL (MSE) & DVFL (SimCLR) & SplitNN \\ \cmidrule(lr){3-3}\cmidrule(lr){4-4}\cmidrule(lr){5-5}
\multicolumn{1}{l|}{\multirow{8}{*}{\rotatebox[origin=c]{90}{\textsc{Quantity}}}} & \multicolumn{1}{c|}{Guest forward (s/ep)}  & 2.144            &   4.872            & 1.830        \\
\multicolumn{1}{c|}{}                          & \multicolumn{1}{c|}{Guest waiting (s/ep)}  &   0.000         &     0         &        2.180\\
\multicolumn{1}{c|}{}                          & \multicolumn{1}{c|}{Guest backward (s/ep)} &   0.650        &     10.488           &         0.001\\
\multicolumn{1}{c|}{}                          & \multicolumn{1}{c|}{Host forward (s/ep)}   &   1.748          &   4.657            &       0.982  \\
\multicolumn{1}{c|}{}                          & \multicolumn{1}{c|}{Host backward (s/ep)}  &   0.806         &    20.778           &       0.274  \\
\multicolumn{1}{c|}{}                          & \multicolumn{1}{c|}{Owner forward (s/ep)}  &   1.271        &    1.149           &        -- \\
\multicolumn{1}{c|}{}                          & \multicolumn{1}{c|}{Owner backward (s/ep)} &   0.564         &   2.430            &         -- \\
\multicolumn{1}{c|}{}                          & \multicolumn{1}{c|}{Guest process (s)}     &  55.883 &      305.596      &            80.220\\
\cline{1-5}
\end{tabular}%
}
\end{table}

The accuracies observed are: 
\begin{itemize}
    \vspace{-2ex}
    \item \textbf{SplitNN} 97.85\%
    \item \textbf{DVFL (MSE)} 97.80\%
    \item \textbf{DVFL (SimCLR)} 97.78\%
\end{itemize}
\vspace*{-2ex}
Accuracies on SimCLR under faults (0 hosts):
\begin{itemize}
    \vspace*{-2ex}
    \item \textbf{Guest}
    \begin{enumerate}
        \item Rejoin Rate = 1.0: 97.63\%
        \item Rejoin Rate = 0.5: 97.64\%
        \item Rejoin Rate = 0.1: 97.41\%
    \end{enumerate}
    \vspace*{-2ex}
    \item \textbf{Communication}
    \begin{enumerate}
        \item Rejoin Rate = 1.0: 97.71\%
        \item Rejoin Rate = 0.5: 97.63\%
        \item Rejoin Rate = 0.1: 97.52\%
    \end{enumerate}
\end{itemize}

The batch size is larger, so losing a single forward pass loses more information.

\end{document}